\documentclass{article} 
\usepackage{iclr2027_conference,times}

\usepackage{amsmath,amsfonts,bm}

\def\eqref#1{equation~\ref{#1}}

\def\1{\bm{1}}

\DeclareMathAlphabet{\mathsfit}{\encodingdefault}{\sfdefault}{m}{sl}
\SetMathAlphabet{\mathsfit}{bold}{\encodingdefault}{\sfdefault}{bx}{n}

\usepackage{hyperref}
\usepackage{url}
\usepackage{cleveref}
\usepackage{booktabs}
\usepackage{graphicx}
\usepackage{makecell}
\usepackage{multirow}
\usepackage{amsfonts}
\usepackage{enumitem}
\usepackage{doi}
\usepackage[T1]{fontenc}

\usepackage[table]{xcolor}
\definecolor{fgreen}{RGB}{34,139,34}
\definecolor{lgreen}{HTML}{6FAF7F} 
\definecolor{fred}{HTML}{B22222} 
\definecolor{lred}{HTML}{D08A8A}
\newcommand{\gain}[1]{{%
  \ifdim #1pt<0.6pt\color{gray}\else\color{fgreen}\fi
  \textbf{\scriptsize\,+#1}}}
\newcommand{\drop}[1]{{%
  \ifdim #1pt<0.6pt\color{gray}\else\color{fred}\fi
  \textbf{\scriptsize\,-#1}}}
\newcommand{\same}{{\color{gray}\textbf{\scriptsize\,±0.0}}}

\newcommand{\gainA}[1]{{%
  \ifdim #1pt<0.6pt\color{fgreen}\else\color{fgreen}\fi
  \textbf{\scriptsize\,+#1}}}
\newcommand{\dropA}[1]{{%
  \ifdim #1pt<0.6pt\color{fred}\else\color{fred}\fi
  \textbf{\scriptsize\,-#1}}}

\usepackage{xfp}
\definecolor{gainC}{HTML}{5DB07A}
\definecolor{dropC}{HTML}{D9706B}
\newcommand{\dcell}[3]{%
  \edef\shadepct{\fpeval{min(75, round(abs(#3) * 55, 0))}}%
  \ifdim #3pt<0pt
    \edef\docolor{\noexpand\cellcolor{dropC!\shadepct}}%
  \else
    \edef\docolor{\noexpand\cellcolor{gainC!\shadepct}}%
  \fi
  \docolor
  #1\textsubscript{\scriptsize$\pm$#2}}
\newcommand{\rcell}[2]{#1\textsubscript{\scriptsize$\pm$#2}}

\title{SAM3RL: Memory Control via Reinforcement Learning for Visual Object Tracking}

\author{Tom\'{a}\v{s} \v{C}\'{i}\v{z}ek\textsuperscript{\rm 1,*,\dag}, Matej Straka\textsuperscript{\rm 1,*}, Illia Volkov\textsuperscript{\rm 2,*}, Mat\v{e}j Kripner\textsuperscript{\rm 1,*}, Ervin Maci\'{c}\textsuperscript{\rm 3},\\
\textbf{Jiri Matas\textsuperscript{\rm 2}, Klara Janouskova\textsuperscript{\rm 2}, Martin Schmid\textsuperscript{\rm 1,4}} \\[2pt]
\textsuperscript{\rm 1}Charles University, Prague, \textsuperscript{\rm 2}Czech Technical University in Prague,\\
\textsuperscript{\rm 3}University of Oxford, \textsuperscript{\rm 4}EquiLibre Technologies, Inc. \\[2pt]
\texttt{\{cizek,straka,schmid\}@kam.mff.cuni.cz, kripner@ufal.mff.cuni.cz}\\
\texttt{\{volkoill,matas,janoukl1\}@fel.cvut.cz, ervin.macic@pmb.ox.ac.uk}\\[2pt]
\textsuperscript{*}Core contributor \quad \textsuperscript{\dag}Project lead
}

\usepackage{xspace}
\makeatletter
\DeclareRobustCommand\onedot{\futurelet\@let@token\@onedot}
\def\@onedot{\ifx\@let@token.\else.\null\fi\xspace}
\def\eg{e.g\onedot,~}   
\def\ie{i.e\onedot,~}

\makeatother

\newif\ifpreprint
\preprinttrue 

\ifpreprint
\iclrfinalcopy
\hypersetup{hidelinks}
\fi

\begin{document}
\maketitle

\ifpreprint
\lhead{SAM3RL: Memory Control via Reinforcement Learning for Visual Object Tracking}
\thispagestyle{plain}
\fi

\begin{abstract}
Segment Anything Model 3 (SAM~3) is a foundation model for image and video segmentation that sets the state of the art in visual object tracking.
Its tracker, inherited from SAM~2, stores information from past frames in a memory bank, managed by a simple First-In-First-Out rule, allowing temporal consistency in complex videos.
Several recent methods improve SAM~2 with hand-crafted memory update rules, each targeting a specific failure mode, such as distractors, occlusions, or object motion. However, their gains transfer only partially to the stronger SAM~3 backbone.
We instead present a more general, data-driven approach that frames memory control as a sequential decision-making task.
SAM3RL is an extension of the SAM~3 tracker with a lightweight memory controller, trained by reinforcement learning, that decides which frames enter the memory bank in order to maximize tracking performance.
SAM3RL sets a new state of the art on 10 of 12 tracking and video object segmentation benchmarks, outperforming all hand-crafted rules without any runtime overhead. 
The same approach applied to SAM~2 gives SAM2RL, which outperforms hand-crafted rules on 9 of the 12 benchmarks, showing the generality of our method across model generations.
\end{abstract}

\section{Introduction}

Visual object tracking is a fundamental computer vision task in which a tracker is given a target object in the first frame of a video and must localize it, as a box or a segmentation mask, in all subsequent frames. Reliable tracking is a key component of applications such as autonomous driving, robotics, video surveillance, and sports analytics~\citep{Marvasti_Zadeh_2022}. The task remains difficult since the target may be surrounded by visually similar distractors, be occluded for long periods, or change its appearance. These challenges are amplified in long videos, where errors can accumulate over time and recovery depends critically on historical context~\citep{javed2021visualobjecttrackingdiscriminative}.

Recent state-of-the-art trackers are built on the \emph{Segment Anything Model} (SAM) family.
SAM~2 \citep{ravi2024sam2segmentimages} is a promptable video segmentation model that conditions each mask prediction on features from past frames held in a small, fixed-size \emph{memory bank}.
SAM~3~\citep{carion2026sam3segmentconcepts} targets promptable concept segmentation, while inheriting the tracker architecture from SAM~2.
The original SAM 2 populates the bank with a simple \emph{First-In-First-Out} (FIFO) rule that keeps the most recent frames regardless of quality or relevance.
Therefore, a line of follow-up work replaces FIFO with hand-crafted update rules, including motion-aware frame selection~\citep{yang2024samuraiadaptingsegmentmodel,chen2025him2samenhancingsam2hierarchical}, distractor-aware memory~\citep{dam4sam_ijcv}, and calibrated prototype memory with positional priors~\citep{xu2025samitepositionpromptedsam2}; some go further and also select among mask candidates by motion consistency~\citep{yang2024samuraiadaptingsegmentmodel,chen2025him2samenhancingsam2hierarchical} or maintain several banks and search over them~\citep{ding2025sam2longenhancingsam2}.
SAM 3 also follows this approach and refines FIFO by keeping only frames in which the object is not occluded, plus the most recent frame.

However, each of the hand-crafted rules targets a particular failure mode, such as distractors, occlusions, or fast motion. Ideally, the tracker should be robust to all of them at the same time. Moreover, these rules are tied to the model generation they were designed for, as all heuristics developed for SAM~2 transfer poorly to the stronger SAM~3. Their gains on tracking benchmarks are limited, and on video object segmentation they even fall below SAM~3 on most benchmarks; see \Cref{fig:radar-baselines}.

\definecolor{oursblue}{HTML}{2B5BB0}
\begin{figure}[t]
  \centering
  \includegraphics[scale=0.95]{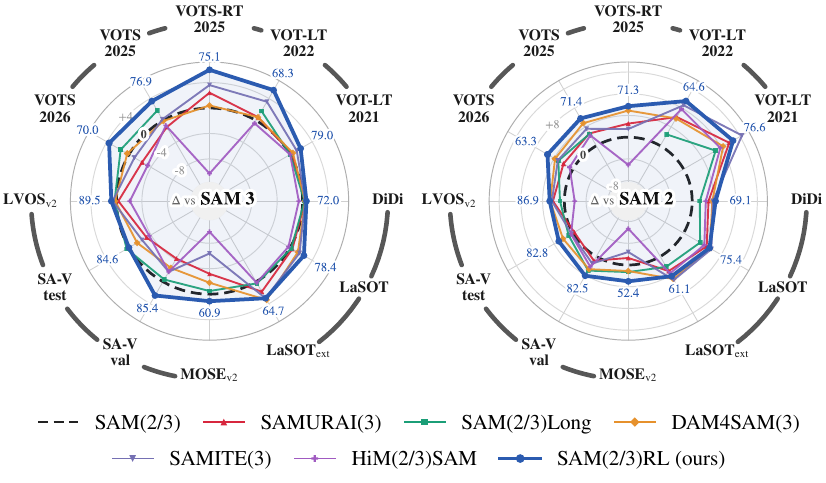}
  \caption{
  \textbf{Gains from hand-crafted rules vs. our learned memory control SAM3RL (left) and SAM2RL (right).}
  Each axis is a tracking or video segmentation benchmark, spanning three categories: VOT(S) challenges (VOTS$_{\text{2025/2026}}$, VOTS-RT$_\text{2025}$, and VOT-LT$_{\text{2021/2022}}$), visual object tracking (DiDi, LaSOT, LaSOT$_{\text{ext}}$), and video object segmentation (MOSE$_{\text{v2}}$, SA-V$_{\text{val/test}}$, LVOS$_{\text{v2}}$). 
  The radius shows the change in that benchmark's standard metric relative to the corresponding baseline tracker (SAM~3 or SAM~2). The \textcolor{oursblue}{outer labels} give our score on each benchmark.
  The compared methods modify what the tracker stores in the memory bank, and in most cases also mask selection or the number of banks. In contrast, our policy operates with the memory bank only. 
  The compared methods are evaluated on SAM~2 as published and on SAM~3 after our porting to that backbone.
  }
  \label{fig:radar-baselines}
\end{figure}

In this work, we show that memory control should be learned rather than designed. We formulate memory control as a \emph{Markov Decision Process} (MDP), designing the state, actions, and reward, but leaving the choice of what to remember to the agent.
That choice is learned by \emph{reinforcement learning} (RL) directly from data, without touching the tracker itself.
We introduce \textbf{SAM3RL}, which applies this memory control formulation to SAM~3 and sets a new state
of the art on several tracking benchmarks, most notably on the VOTS and VOT-LT challenges.
It does so at no additional inference cost and, unlike hand-crafted rules, without sacrificing performance on video object segmentation.
Applied to SAM~2, the same recipe yields SAM2RL, which still outperforms hand-crafted rules on most benchmarks, showing the generality of our learned method across model generations.

Our contributions are as follows:

\noindent\textbf{Learned memory control for SAM-based trackers.}
We formulate memory bank control in SAM-based trackers as an MDP over a fixed-capacity bank. 
The state describes the content of the bank, the action selects the entry to evict,
and the reward scores the resulting prediction against the annotation. 
This formulation allows the memory control to be learned instead of being manually designed.

\noindent\textbf{State-of-the-art visual object tracker.}
We present SAM3RL, which extends SAM~3 with a memory controller obtained by optimizing the introduced MDP with RL.
It improves over SAM~3 on 11 out of 12 benchmarks and ranks first on 10 of them.
The largest gains are achieved on the VOT(S) challenges, with $+3.6$ on VOTS$_{\text{2026}}$ and $+5.5$ on VOT-LT$_{\text{2022}}$. 
It is also the only tracker that consistently succeeds in video object segmentation, scoring $+1.1$ on MOSE$_{\text{v2}}$ and $+2.5$ on~SA-V$_{\text{val}}$.
SAM3RL achieves this while optimizing the memory control only, whereas most compared methods additionally modify mask selection, prompting, or the number of banks maintained.

\noindent\textbf{Online controller without a runtime penalty.}
The SAM3RL controller is a lightweight MLP evaluated once per memory update.
It reads only values the tracker has already computed, and stays fully online without lookahead or extra feature extraction, unlike other methods.
On tracking benchmarks, SAM3RL is faster than SAM~3, at 16.0\,FPS against 15.3\,FPS, as the memory is updated only every sixth frame, matching the decision rate used in training.
Under the real-time FPS constraint of VOTS-RT$_{\text{2025}}$, SAM3RL stays best at $75.1$, dropping only $1.8$ points, while SAM~3 drops $4.3$.

\noindent\textbf{Generality across model generations.}
SAM2RL is obtained by applying the same recipe to SAM~2.
It improves the baseline on all 12 benchmarks and leads on 9 of them, although every compared rule was designed specifically for SAM~2.
SAM2RL still remains faster than SAM~2, at 25.5\,FPS against 25.1\,FPS.
Together with SAM3RL, this demonstrates the generality of the introduced method.
\section{Related Work}
\label{sec:related}

\emph{Visual object tracking} (VOT) and \emph{semi-supervised video object segmentation} (VOS) are conditioned on a target specified in the first frame of a video, which must then be localized in all subsequent frames.
The target is represented by a bounding box in VOT and a pixel-level mask in VOS.
Both tasks require maintaining the target through appearance changes, occlusions,
fast motion, and distractors, which remain the dominant sources of tracking failure
\citep{kristan2020vot}. To address these challenges, trackers have long exploited information from previous frames~\citep{1288530,6104061,oh2019videoobjectsegmentationusing,cui2022mixformerendtoendtrackingiterative,han2024actrackaddingspatiotemporalcondition}.

Recently, the SAM~2 foundation model \citep{ravi2024sam2segmentimages} set a new state of the art in tracking.
Its impact is evident in the VOTS$_{\text{2026}}$ challenge, where 92\% of submitted trackers were derived from the SAM family~\citep{kristan2026the}.
Central to the SAM~2 design is the memory bank, a fixed-capacity store of past frames on which every prediction is conditioned by memory attention.
The bank holds seven slots, the prompted first frame and the six most recent ones, under a FIFO rule.
Because adjacent frames are highly correlated, FIFO tends to fill the bank with redundant observations rather than the diverse history required to distinguish the target from distractors or recover after occlusions.

Consequently, memory design in SAM-based trackers has become an active research area, with follow-up work replacing FIFO with hand-crafted selection rules.
SAMURAI \citep{yang2024samuraiadaptingsegmentmodel} uses a Kalman filter \citep{10.1115/1.3662552} to score motion
consistency, both selecting among per-frame mask candidates and gating memory updates.
DAM4SAM \citep{dam4sam_ijcv} splits the bank into a fixed-interval part and a
part reserved for frames that resolve visually similar distractors. SAMITE
\citep{xu2025samitepositionpromptedsam2} calibrates stored prototypes and adds positional priors.
HiM2SAM~\citep{chen2025him2samenhancingsam2hierarchical} combines hierarchical motion estimation with a
long- and short-term split of the bank.
SAM2Long \citep{ding2025sam2longenhancingsam2} keeps several banks in parallel and performs tree search over alternative trajectories. Even SAM~3~\citep{carion2026sam3segmentconcepts} relies on such a rule, retaining an unprompted frame only when the model is confident that the target is present.

However, for older trackers, learned memory control has already been shown to outperform hand-crafted rules~\citep{choi2018realtimevisualtrackingdeep,yang2019visualtrackingdynamicmemory,zhang2019learningmodelupdatesiamese,9156764}.
\ifpreprint
Specifically for SAM~2, we attempted learning the rules instead of designing them in our preliminary work \citep{adamyan2025sam2rlreinforcementlearningmemory}.
There, we showed the potential of learned memory control by fitting a separate RL agent to each of a handful of training videos.
Agents improve over the hand-crafted rules, but do not transfer to unseen videos.
This demonstrates a performance upper bound rather than a usable tracker.
Here, we develop the full method, a single general-purpose policy that generalizes to unseen videos, extends to SAM~3, and sets a new state of the art in visual object tracking.
\fi
\begin{figure}[t]
\centering
\includegraphics[width=1\textwidth]{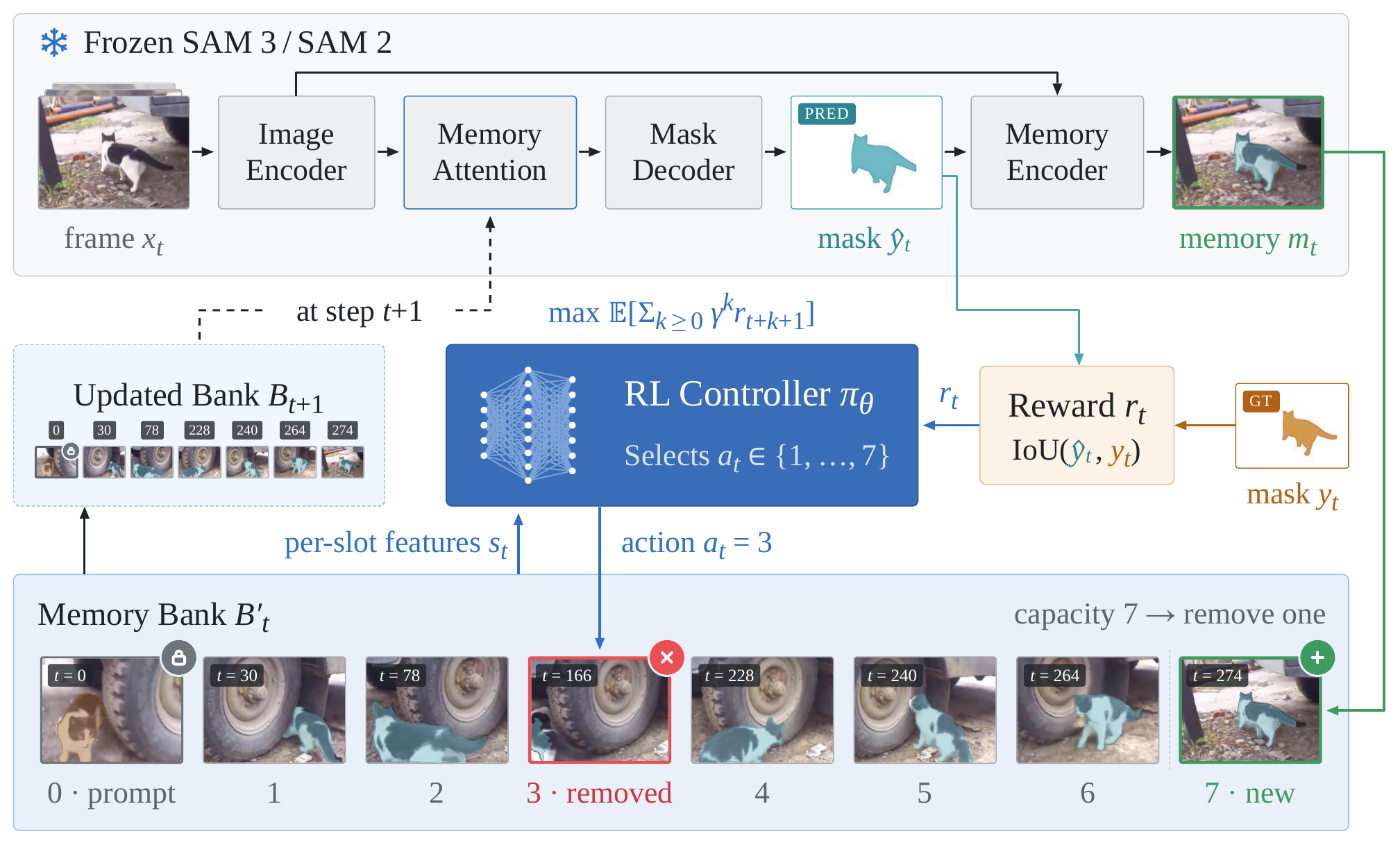}
\caption{\textbf{Learned memory control for SAM-based trackers.}
  The frozen tracker (top) embeds the frame $x_t$ and conditions it through memory attention on the current memory bank $B_t$.
  The decoder predicts the mask $\hat{y}_t$, and the memory encoder fuses it with $x_t$ into a new memory $m_t$.  
  The bank has capacity seven, where slot $0$ is permanently reserved for the conditioning frame.
  Appending $m_t$ to slot $7$ therefore overfills the bank, so one of the slots $1, \dots, 7$ must be evicted.
  Our controller $\pi_\theta$ observes the state $s_t$, which collects temporal, confidence, semantic, and geometric features of all eight slots, and picks the action $a_t \in \{1, \dots, 7\}$ that removes the corresponding memory (here $a_t = 3$).
  The surviving slots form the updated bank $B_{t+1}$ that the tracker reads at step $t+1$.
  During training, $\pi_\theta$ receives the reward $r_t$, scoring $\hat{y}_t$ against the annotation $y_t$, and maximizes the expected discounted return $\mathbb{E}_{\pi_\theta}\!\big[\sum_{k\ge0}\gamma^k r_{t+k+1}\big]$.
  At inference, $\pi_\theta$ is fixed and the actions are taken greedily.
  }
\label{fig:sam3rl}
\end{figure}

\section{Methodology}
\label{sec:method}
We formulate memory control as an MDP and train a lightweight MLP controller $\pi_\theta$ with RL by rolling out tracking episodes over a corpus of annotated videos.
Given a training sequence with frames $x_0, \dots, x_T$ and object mask or box annotations $y_0, \dots, y_T$, the simulation starts by conditioning the frozen tracker on the first annotation $y_0$; see Figure~\ref{fig:sam3rl}.
At each step $t$, the tracker holds a memory bank $B_t = (m_{\tau_0}, m_{\tau_1}, \dots, m_{\tau_6})$ of seven memory frames with indices $0 = \tau_0 < \tau_1 < \cdots < \tau_6 < t$, where slot $0$ is permanently occupied by the conditioning frame $m_0$ and the other six hold past memory frames formed from the tracker's own predictions.
The image encoder embeds $x_t$ and the memory attention conditions this embedding on the current content of $B_t$. 
The mask decoder then converts the result into three candidate masks and keeps the most confident as the prediction $\hat{y}_t$.
The mask decoder also outputs an object pointer $p_t$, which is a single token summarizing the appearance of the target.
The memory encoder then fuses $\hat{y}_t$ with the embedding of $x_t$ into a spatial memory feature $f_t$, and the two together form the memory frame $m_t = (p_t, f_t)$.

We assume that the bank $B_t$ is always full, except during the first six frames $x_1, \dots, x_6$, when the memory frames $m_1, \dots, m_6$ are simply appended without eviction. Once $B_t$ is filled, storing $m_t$ requires evicting one of the existing entries.
Whereas prior work relies on hand-crafted rules, we employ a learned controller $\pi_\theta$ that selects one of seven candidates for removal, either one of the stored memories $m_{\tau_1}, \dots, m_{\tau_6}$ or the newly formed memory $m_t$.
The six remaining candidates, together with the conditioning memory $m_0$, form the updated bank $B_{t+1}$ used for the tracker prediction in the next step.
Once training is complete, the controller $\pi_\theta$ is fixed and the same loop is used for inference on unseen videos, with actions taken greedily and no reward computed.

\subsection{MDP Formulation}
\label{sec:mdp}

An \emph{episode} of our MDP is one object tracked through a video.
The policy $\pi_\theta$ acts once per tracker step, after $\hat{y}_t$ and $m_t$ have been produced and before $B_t$ is updated.
We denote a temporary bank after appending the new memory frame $m_t$ by $B'_t = (m_{\tau_0}, \dots, m_{\tau_7})$ where $\tau_7 = t$.
The policy observes the state $s_t$ and selects the action $a_t$ that removes one of $m_{\tau_1}, \dots, m_{\tau_7}$, producing the next memory bank $B_{t+1}$.
It also receives the reward $r_t$, which reflects the quality of the prediction $\hat{y}_t$ against the annotation $y_t$.
We now formalize the state $s_t$, action $a_t$, and reward $r_t$.

\noindent\textbf{State space.}
The state $s_t$ describes the content of $B'_t$ by four pairwise disjoint groups of features, computed for every slot $i \in \{ 0, \dots, 7 \}$ from the tracker outputs at the corresponding time step $\tau_i$:

\begin{itemize}[leftmargin=1em, itemsep=2pt, topsep=2pt, parsep=0pt, partopsep=0pt]
  \item \emph{Temporal encodings} are sinusoidal positional encodings
        $\mathrm{PE}_{\tau_i} \in \mathbb{R}^{32}$ of the frame age $t - \tau_i$.

  \item \emph{Confidence features} are the object score $o_{\tau_i} \in \mathbb{R}$,
        which is the logit of the target being visible in $x_{\tau_i}$, and the predicted IoU
        scores $\iota_{\tau_i} \in [0,1]^3$, one per candidate mask, estimating how
        well that mask overlaps with the ground truth. Both $o_{\tau_i}$ and $\iota_{\tau_i}$ are outputs of the tracker's mask decoder.

  \item \emph{Semantic features} are the object pointer $p_{\tau_i} \in \mathbb{R}^{256}$
        and the mean-pooled memory feature $f^{pool}_{\tau_i} \in \mathbb{R}^{64}$ of the spatial memory feature $f_{\tau_i}$ under the predicted mask $\hat{y}_{\tau_i}$ to reduce the input size. 
        The features also contain pairwise cosine similarities between $(p_{\tau_i}, p_{\tau_j})$ and
        $(f^{pool}_{\tau_i}, f^{pool}_{\tau_j})$ for $i \neq j$.

  \item \emph{Geometric features} describe the area, centroid, and box size of
        $\hat{y}_{\tau_i}$, plus the pairwise mask IoUs between $\hat{y}_{\tau_i}$ and
        $\hat{y}_{\tau_j}$ for $i \neq j$. They further contain the IoUs of the two
        unselected candidate masks for $\hat{y}_t$ against every $\hat{y}_{\tau_i}$, as
        well as the IoU between the two unselected masks themselves.
\end{itemize}

\noindent\textbf{Action space.}
The action $a_t \in \{1, \dots, 7\}$ is the index of the memory frame
that the policy removes from $B'_t$, giving the updated bank $B_{t+1} = B'_t \setminus \{m_{\tau_{a_t}}\}$.
Choosing $a_t = 7$ removes the newly formed $m_t$ and thus rejects the current frame, leaving the bank unchanged at $B_{t+1} = B_t$.
The actions also determine which object pointers the tracker attends to.
In fact, memory attention reads sixteen object pointers but only seven memory features.
We take the seven memory features and object pointers from $B_t$ and fill the remaining nine pointers from the frames most recently evicted from the bank.
We keep temporal encodings for the memory attention inputs from SAM~3 but index object pointers by recency rank instead of frame offset, so their encodings stay within the range seen in training.

The policy then defines a distribution $\pi_\theta(a_t \mid s_t)$ over the seven actions, from which we sample during training and take the most likely action during inference.
The memory rules of SAM~2 and SAM~3 can themselves be formulated as policies in this action space.
SAM~2 always selects $a_t = 1$ and evicts the oldest memory, while SAM~3 keeps the most recent frame and selects $a_t = 6$ whenever an object-score-gated IoU of $x_{\tau_6}$ falls below a predefined threshold, and $a_t = 1$ otherwise.

\noindent\textbf{Reward.}
The reward measures the tracking quality of the prediction $\hat{y}_t$ against the annotation $y_t$ by
\begin{equation}
q_t =
\begin{cases}
1 & \text{if }  \ \hat{y}_t = \emptyset \text{ and } \ y_t = \emptyset, \\[2pt]
\mathrm{IoU}(\hat{y}_t, y_t) & \text{otherwise},
\end{cases}
\end{equation}
which credits the tracker for the overlap between $\hat{y}_t$ and $y_t$ when the target is visible, and for correctly reporting when the object is absent.
If the annotation is a bounding box, the IoU is computed against the tight
axis-aligned box enclosing the predicted mask.
A large amount of variance of $q_t$ comes from the difficulty of the video rather than from the actions taken.
Therefore, we normalize the reward by setting $r_t = q_t - b_t$, where $b_t$ is the tracking quality that the underlying tracker scores at the same frame of the same video under the same conditioning with SAM 3's default memory control.
The baseline depends only on the video and not on the policy $\pi_\theta$, so we pre-compute $b_t$ only once for the whole training dataset.
The policy then maximizes the expected return $\mathbb{E}_{\pi_\theta}\!\left[G_t\right]$,
where $G_t = \sum_{k\ge0}\gamma^k r_{t+k+1}$ is the discounted return from step $t$ and $\gamma$ is the discount factor.

\Cref{suppl:ablations} ablates the input feature groups, object pointer selection, and the reward normalization.

\subsection{Training the Memory Controller}
\label{sec:learning}

\noindent\textbf{Architecture.}
The controller $\pi_\theta$ is a single MLP, which serves as the \emph{actor}.
All features enter it directly, except the object pointers, which a shared linear projection first compresses to 64 dimensions.
All groups are then concatenated and passed through one hidden layer of 2{,}048 units and a linear head over the seven actions, about 3.2\,M parameters in total.
For training, we add a \emph{critic}, a second network of the same design with a scalar head, which estimates the expected return, shares no parameters with the actor, and is discarded after training.
Besides $s_t$, the critic receives the normalized number of decision steps left in the episode, which it needs to predict undiscounted finite-horizon returns.
This value is unavailable at inference, but since only the actor is deployed, the controller remains fully online and depends on $s_t$ alone.

\noindent\textbf{Training data.}
We train on the LaSOT training split \citep{fan2019lasothighqualitybenchmarklargescale}, whose 1{,}120 sequences are divided into 896 for training and 224 for validation, while the official test set is held out.
LaSOT is annotated with per-frame bounding boxes at 30\,FPS.
Since consecutive frames at that rate are nearly identical, an eviction decision changes the bank very little while still consuming a decision step.
We therefore subsample every sixth frame, so that training episodes run at 5 FPS and span 425 decision steps on average.
This also matches the typical annotation rate of VOS datasets.

\noindent\textbf{Optimization.}
We train the controller with Proximal Policy Optimization (PPO) \citep{schulman2017proximalpolicyoptimizationalgorithms}, using the undiscounted Monte-Carlo return (\ie $\gamma = 1$) minus the critic's predicted value as the advantage.
The return is bootstrapped only when the rollout window truncates an episode, corresponding to $\lambda = 1$ in generalized advantage estimation (GAE)~\citep{schulman2018highdimensionalcontinuouscontrolusing}.
Apart from this bootstrap, we keep the critic out of the targets and use it only as a baseline, as it fits the training videos well but transfers poorly to unseen videos.
Each of the 400 iterations collects 376-step rollouts from 256 parallel environments.
Training takes about 24 hours on four H200 GPUs.
We evaluate the greedy policy on the full validation videos every 25 iterations and select the best checkpoint.
All remaining hyperparameters are listed in~\Cref{tab:supp_hparams} in \Cref{suppl-hyperparameters}.

\noindent\textbf{Inference.}
The actor picks actions greedily, and its only overhead is a single MLP evaluation per memory update.
To keep the decision rate close to the 5\,FPS seen in training, the memory controller operates with a stride $s = \text{round}(\nu / 5)$, where $\nu$ is the frame rate of the benchmark.
The tracker still predicts a mask on every frame, but only every $s$th frame triggers a memory update and can enter the bank.
On the remaining frames no memory frame $m_t$ is formed, so the memory encoder that produces $f_t$ is skipped altogether.
The 30\,FPS VOT and VOTS benchmarks use $s = 6$, while the 5\,FPS MOSE$_{\text{v}2}$ uses $s = 1$.
The frame rates and strides for all benchmarks are listed in \Cref{suppl-eval}.
\section{SAM3RL Performance}
\label{sec:results}

SAM3RL sets a new state of the art among memory control methods for SAM-based trackers on most tracking and video object segmentation benchmarks.
In this section, we compare it with the prior methods discussed in \Cref{sec:related}, all evaluated on the SAM~3 backbone.
Because these methods were designed for SAM~2 and their authors have not released SAM~3 implementations, we port each one to SAM~3 ourselves.
Each port replaces the SAM~2 video predictor with the SAM~3 tracker and leaves the method otherwise untouched.
We also compare the prior methods, in their original SAM~2 implementations, against SAM2RL (our memory control on SAM~2) in \Cref{fig:radar-baselines} and \Cref{suppl-results}.
\Cref{suppl-eval} details tracker evaluation, FPS measurement, reproducibility, and the ports.

\subsection{VOT(S) Challenges}
\label{sec:vot-competitions}

The VOT initiative has organized annual tracking challenges for over a decade, and its benchmarks have evolved with the field~\citep{kristan2013}.
Current VOTS challenges unify short-term and long-term tracking with single-target and multi-target scenarios.
The sequences feature occlusions, visually similar distractors, and targets that leave the field of view and re-enter.
VOTS targets are annotated by segmentation masks and scored on a remote server, whereas VOT-LT targets are annotated by bounding boxes.
Results are presented in~\Cref{tab:vot_competitions}.
The challenges separate the methods most sharply, with scores spanning $7.0$ points on VOTS$_{\text{2026}}$.
SAM3RL ranks first on all five challenges.
Among hand-crafted rules, SAMITE3 and SAM3Long provide the strongest baselines.

\textbf{VOTS$_{\text{2026}}$}~\citep{kristan2026the} is the latest edition of the challenge, and \textbf{VOTS$_{\text{2025}}$}~\citep{10350713} is the previous one, whose sequences are a subset of it. 
\textbf{VOTS-RT$_{\text{2025}}$}~\citep{11375669} is the real-time variant of the
VOTS$_{\text{2025}}$ benchmark, designed to evaluate tracking quality under a strict
streaming constraint. Frames are delivered
to the tracker at a constant rate of 20\,FPS, independently of its processing
latency. The tracker must finish reporting the mask in the
current frame before the next frame becomes available. Otherwise, the prediction from the previous frame is retained. The constraint costs SAM~3 several points, whereas SAM3RL gains the most of all methods. Notably, we do not report SAM3Long in this setting because it is not an online
tracker and cannot satisfy this causal protocol; see \Cref{suppl-eval}.
\textbf{VOT-LT$_{\text{2021}}$}~\citep{9607833} evaluates long-term tracking, where the
target is absent for extended periods and must be redetected, and \textbf{VOT-LT$_{\text{2022}}$}~\citep{10.1007/978-3-031-25085-9_25} is the latest LT challenge, which introduced more difficult sequences.
 The results on VOT-LT$_{\text{2021}}$ were the most affected by a race condition in the original SAM~3 video predictor, which we identify and fix in \Cref{suppl-race-condition}; the numbers reported in this work are measured after the fix.

\textbf{Model throughput} is measured on VOTS$_{2026}$ and covers model inference only.
On the 30\,FPS benchmarks, SAM3RL updates its memory only every $s$th frame with $s=6$, matching the 5\,FPS data used during training.
On the $s - 1$ frames in between, no memory frame is formed, so the memory encoder is skipped and the controller is not evaluated.
SAM~3, in contrast, runs the memory encoder on every frame.
The time saved on memory encoding outweighs the cost of the controller, so SAM3RL reaches 16.0~FPS, exceeding both the hand-crafted rules and the 15.3~FPS of SAM~3.
The speedup therefore comes from the less frequent memory updates rather than from the policy itself.
With memory updated on every frame where $s = 1$, SAM3RL runs at 14.9~FPS; see Table~\ref{tab:sam3_stride_ablation}.

\subsection{VOT Datasets}

The VOT datasets are annotated with bounding boxes, so each predicted mask is scored by its tight box;
results are in~\Cref{tab:vot_datasets}.
SAM3RL improves over SAM~3 on all three and ranks first on two of them.
DAM4SAM3 and SAMITE3 are the strongest rules here, while HiM3SAM is the weakest.

\begin{table*}[t]
\setlength{\tabcolsep}{0.5em}
\caption{\textbf{Performance on VOT(S) challenges.} We report standard metrics for the evaluated challenges. Deltas indicate improvement or decline relative to SAM 3. \textbf{Bold} indicates the best.}
\label{tab:vot_competitions}
\begin{center}
\begin{tabular}{l c c c c c c}
\toprule
& & \multicolumn{3}{c}{Quality} & \multicolumn{2}{c}{F-score} \\
\cmidrule(lr){3-5} \cmidrule(lr){6-7}
Model & FPS & VOTS$_{\text{2026}}$ & VOTS$_{\text{2025}}$ & VOTS-RT$_{\text{2025}}$ & VOT-LT$_{\text{2021}}$ & VOT-LT$_{\text{2022}}$ \\
\midrule
SAM 3      & 15.3 & 66.4 & 73.4 & 69.1 & 77.1 & 62.8 \\ \midrule
SAMURAI3    & 15.1 & 64.0\drop{2.4} & 72.2\drop{1.2} & 71.5\gain{2.4} & 77.4\gain{0.3} & 63.4\gain{0.6} \\
SAM3Long   & 9.3& 67.9\gain{1.5} & 75.2\gain{1.8} & -- & 77.1\same & 64.5\gain{1.7} \\
DAM4SAM3   & 14.2& 66.7\gain{0.3} & 73.3\drop{0.1} & 69.5\gain{0.4} & 77.6\gain{0.5} & 63.4\gain{0.6} \\
SAMITE3    & 12.7& 65.4\drop{1.0} & 73.6\gain{0.2} & 72.7\gain{3.6} & 78.3\gain{1.2} & 66.2\gain{3.4} \\
HiM3SAM    & 11.8& 63.0\drop{3.4} & 72.5\drop{0.9} & 58.8\drop{10.3} & 77.1\same & 62.3\drop{0.5} \\ \midrule
\textbf{SAM3RL (ours)} & \textbf{16.0}& \textbf{70.0}\gain{3.6} & \textbf{76.9}\gain{3.5} & \textbf{75.1}\gain{6.0} & \textbf{79.0}\gain{1.9} & \textbf{68.3}\gain{5.5} \\
\bottomrule
\end{tabular}
\end{center}
\end{table*}
\begin{table*}[t]
\setlength{\tabcolsep}{0.4em}
\begin{minipage}[t]{0.44\textwidth}
\caption{\textbf{Performance on VOT datasets.} Deltas indicate improvement or decline relative to SAM 3. \textbf{Bold} indicates the best.}
\label{tab:vot_datasets}
\begin{center}
\small
\begin{tabular}{l c c c }
\toprule
& Quality\vphantom{$\mathcal{J\&\dot{F}}$} & \multicolumn{2}{c}{AUC}\\
\cmidrule(lr){2-2} \cmidrule(lr){3-4} 
Model & DiDi & LaSOT & LaSOT$_{\text{ext}}$ \\
\midrule
SAM 3   & 71.4 & 75.9 & 61.7\\ \midrule
SAMURAI3   & 71.8\gain{0.4} & 76.3\gain{0.4} & 63.5\gain{1.8} \\
SAM3Long   & 71.7\gain{0.3} & 76.1\gain{0.2} & 62.0\gain{0.3} \\
DAM4SAM3   & 71.8\gain{0.4} & 77.3\gain{1.4} & \textbf{65.0}\gain{3.3} \\
SAMITE3    & 71.8\gain{0.4} & 77.6\gain{1.7} & 64.7\gain{3.0} \\
HiM3SAM    & 70.8\drop{0.6} & 75.6\drop{0.3} & 61.9\gain{0.2} \\ \midrule
\textbf{SAM3RL  } & \textbf{72.0}\gain{0.6} & \textbf{78.4}\gain{2.5} & 64.7\gain{3.0} \\
\bottomrule
\end{tabular}
\end{center}
\end{minipage}
\hfill
\begin{minipage}[t]{0.53\textwidth}
\caption{\textbf{Performance on VOS datasets.} Deltas indicate improvement or decline relative to SAM 3. \textbf{Bold} indicates the best.
}
\label{tab:vos_results}
\begin{center}
\small
\begin{tabular}{l c c c c }
\toprule
& \multicolumn{1}{c}{$\mathcal{J\&\dot{F}}$} & \multicolumn{3}{c}{$\mathcal{J\&F}$} \\
\cmidrule(lr){2-2} \cmidrule(lr){3-5}
Model & MOSE$_{\text{v2}}$ & SA-V$_{\text{val}}$ & SA-V$_{\text{test}}$ & LVOS$_{\text{v2}}$ \\
\midrule
SAM 3   & 59.8& 82.9& 84.6& 88.7\\ \midrule
SAMURAI3  & 56.6\drop{3.2}& 78.7\drop{4.2}& 81.4\drop{3.2}& 88.4\drop{0.3} \\
SAM3Long  & 59.3\drop{0.5}& 82.5\drop{0.4}& \textbf{85.0}\gain{0.4}& 89.0\gain{0.3} \\
DAM4SAM3  & 58.0\drop{1.8}& 80.4\drop{2.5}& 83.1\drop{1.5}& 88.9\gain{0.2}\\
SAMITE3    & 53.4\drop{6.4}& 81.1\drop{1.8}& 82.2\drop{2.4}& 89.3\gain{0.6} \\
HiM3SAM    & 50.0\drop{9.8}& 81.1\drop{1.8}& 80.9\drop{3.7}& 86.6\drop{2.1} \\ \midrule
\textbf{SAM3RL} & \textbf{60.9}\gain{1.1} & \textbf{85.4}\gain{2.5} & 84.6\same & \textbf{89.5\gain{0.8}} \\  
\bottomrule
\end{tabular}
\end{center}
\end{minipage}
\end{table*}

\textbf{DiDi}~\citep{dam4sam_ijcv} is a set of distractor-distilled sequences from various tracking datasets, built to isolate the failures caused by visually similar objects.
It was introduced with DAM4SAM, which was designed specifically against distractors.
On SAM~2, DAM4SAM improves clearly on DiDi (\Cref{suppl-results}), but on SAM~3 most of the gain disappears.
\textbf{LaSOT}~\citep{fan2019lasothighqualitybenchmarklargescale} is the standard long-term tracking benchmark.
Our policy is trained on the LaSOT training split, so its test set is the closest to our training distribution among all benchmarks.
\textbf{LaSOT$_{\text{ext}}$}~\citep{fan2020lasothighqualitylargescalesingle} extends LaSOT with sequences of object classes not present in the LaSOT training set.
It is the only tracking benchmark in the main comparison on which SAM3RL does not rank first.
The gap to the best method is small, and the three leading methods stay well ahead of the remaining rules.

\subsection{VOS Datasets}
\label{sec:vos}

The VOS datasets are annotated with segmentation masks, so predictions are scored at the pixel level; results are in~\Cref{tab:vos_results}.
SAM3RL improves over the baseline on three of the four benchmarks and ranks first on them.
The rules that help most on tracking hurt segmentation, whereas our learned memory controller, trained only on box-annotated LaSOT, remains effective in both settings.

\textbf{MOSE$_{\text{v2}}$}~\citep{ding2025mosev2challengingdatasetvideo} targets conditions largely absent from the classic VOS benchmarks, including frequent disappearance and reappearance, severe occlusions and crowding, small objects, and adverse lighting.
It is the hardest VOS benchmark in our evaluation.
All hand-crafted rules fall below the baseline, and SAM3RL is the only method to improve on it.
\textbf{LVOS$_{\text{v2}}$}~\citep{hong2024lvosbenchmarklargescalelongterm} evaluates long-term segmentation, where the memory bank must retain information over far longer horizons than on the other VOS benchmarks.
It is the only VOS benchmark on which most rules improve over the baseline.
\textbf{SA-V}~\citep{ravi2024sam2segmentimages} was released with SAM~2, covering diverse everyday objects.
SAM3RL leads on the validation split and matches the baseline on the test split.
\begin{table*}[t]
\setlength{\tabcolsep}{0.3em}
\caption{\textbf{Memory stride ablation.}
All benchmarks here run at 30\,FPS, so $s = 6$ matches the 5\,FPS training and $s = 1$ updates every frame.
SAM~3 at $s=1$ is the original tracker.
SAM~3 at $s=6$ is the official strided variant of SAM~3, which encodes memory and reselects the bank every frame $t$, keeping $t-1$ frame plus the frames $t-1-s, t-1-2s, \dots$ that pass SAM~3's threshold.
SAM3RL and SAM~3$_{\text{ours}}$ update the bank only every $s$th frame and hold it fixed in between, with SAM~3$_{\text{ours}}$ using SAM~3's selection rule.
Deltas are relative to SAM~3 at $s=1$. \textbf{Bold} indicates the best.
}
\label{tab:sam3_stride_ablation}
\begin{center}
\small
\begin{tabular}{l @{\hspace{0.5em}} c @{\hspace{1.5em}} c @{\hspace{1.5em}} c c c c c c c}
\toprule
& & & \multicolumn{3}{c}{Quality} & \multicolumn{2}{c}{F-score} & \multicolumn{2}{c}{AUC} \\
\cmidrule(lr){4-6} \cmidrule(lr){7-8} \cmidrule(lr){9-10}
Model & $s$ & FPS & VOTS$_{\text{2026}}$ & VOTS$_{\text{2025}}$ & DiDi & VOT-LT$_{\text{2021}}$ & VOT-LT$_{\text{2022}}$ & LaSOT & LaSOT$_{\text{ext}}$ \\
\midrule
SAM 3                   & $1$ & 15.3 & 66.4 & 73.4 & 71.4 & 77.1 & 62.8 & 75.9 & 61.7 \\
\textbf{SAM3RL} & $1$ & 	14.9 & 69.3\gain{2.9} & 75.4\gain{2.0} & 71.7\gain{0.3} & 77.2\gain{0.1} & 65.6\gain{2.8} & 77.1\gain{1.2} & \textbf{65.2}\gain{3.5} \\
\midrule
SAM 3   & $6$ & 15.5 & 55.5\drop{10.9} & 66.9\drop{6.5} & 60.5\drop{10.9} & 76.2\drop{0.9} & 61.8\drop{1.0} & 72.0\drop{3.9} & 57.4\drop{4.3} \\
SAM 3$_{\text{ours}}$                  & $6$ & 15.8 & 68.1\gain{1.7} & 75.9\gain{2.5} & 71.2\drop{0.2} & 75.7\drop{1.4} & 65.6\gain{2.8} & 77.4\gain{1.5} & 64.8\gain{3.1} \\
\textbf{SAM3RL} & $6$ & \textbf{16.0} & \textbf{70.0}\gain{3.6} & \textbf{76.9}\gain{3.5} & \textbf{72.0}\gain{0.6} & \textbf{79.0}\gain{1.9} & \textbf{68.3}\gain{5.5} & \textbf{78.4}\gain{2.5} & 64.7\gain{3.0} \\
\bottomrule
\end{tabular}
\end{center}
\end{table*}

\section{Analysis of Learned Memory Control}
\label{sec:analysis}

\noindent\textbf{Memory stride.}
Deciding on every frame is the most flexible choice in principle, but training at 30\,FPS is costly, as most decisions concern nearly identical frames.
We therefore train at 5\,FPS and match that decision rate at inference with the stride $s$.
\Cref{tab:sam3_stride_ablation} shows the effect of stride on 30\,FPS benchmarks.
At $s = 1$, our policy runs outside its training distribution, yet still outperforms SAM~3 on all seven benchmarks, by 1.8 points on average.
At $s = 6$, although only every sixth frame can enter the bank, the gain grows to 2.9 points as the decision rate matches the 5\,FPS used in training.
In our evaluations, the compared methods run at their native strides, as most were developed for 30\,FPS benchmarks, and DAM4SAM and HiM2SAM even adapt their stride to the benchmark frame rate, similarly to ours.
The SAM~3 paper, however, reports only VOS results at $s = 1$, so we also evaluate SAM~3 at $s = 6$, both with its official strided variant and as our modification SAM~3$_\text{ours}$, which pairs SAM~3's selection rule with our stride.
The former loses up to 10.9 points, while the latter gains 1.4 points on average.
At equal stride, SAM3RL still outperforms SAM~3$_\text{ours}$ by 1.5 points on average.

\begin{figure}[!htb]
  \centering
  \includegraphics[scale=0.985]{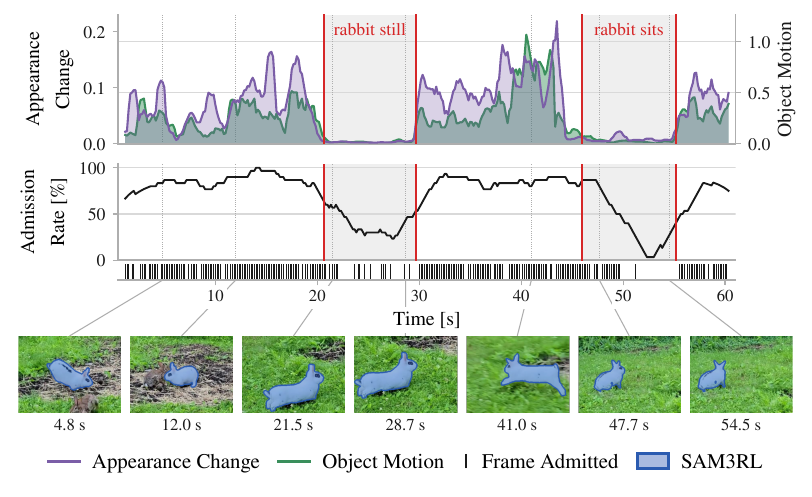}
  \vspace{-1.5em}
  \caption{\textbf{Admission rate of incoming frames by SAM3RL follows object change.}
  An LVOS$_{\text{v2}}$ video sequence in which a rabbit settles and hops off twice over 60\,s.
  \emph{Object motion} is the shift of the ground-truth box center between consecutive decisions, normalized by the square root of the box area.
  \emph{Appearance change} is the DINOv3 \citep{simeoni2025dinov3} cosine distance between consecutive ground-truth crops.
  The middle panel is the admission rate in a 5\,s window, one tick per admitted decision.
  Admissions are sparse while the rabbit stays still (gray) and dense when both curves rise.
}  
  \label{fig:rabbit}
\end{figure}

\noindent\textbf{Learned memory selection.}
SAM~3 admits only frames whose effective IoU score exceeds a threshold.
Its bank thus acts as FIFO while the target is visible and, during absences, holds the pre-absence frames plus the most recent one.
\Cref{fig:age} shows that SAM3RL instead spreads its bank further back in time and, during absences, mixes older pre-absence frames with ones from the absence itself.
Unlike SAM~3, the policy admits nearly all medium-score frames and grows more selective as the score rises, rejecting the most confident frames about as often as those predicted absent; see \Cref{fig:admission}.
The policy also adapts to video content, admitting far more frames when the target moves or changes appearance, as in \Cref{fig:rabbit}.
\Cref{suppl:analysis} quantifies this effect and includes more behavior analysis.

\ifpreprint
\begin{figure*}[!htb]
\begin{minipage}[t]{0.49\textwidth}
\centering
\includegraphics[scale=0.93]{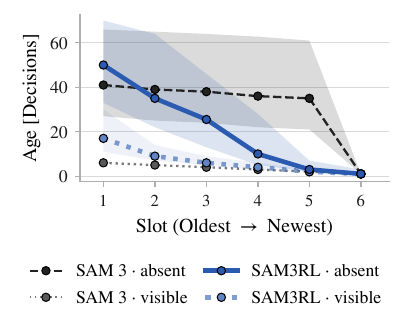}
\vspace{-1.5em}
\caption{\textbf{Frame age in memory slots} for SAM3RL and SAM~3 on MOSEv2.
\emph{Visible} pools all
decisions with the target present.
\emph{Absent} pools the last decisions of
absent gaps of at least 16 decisions.
Median with the 25th to 75th percentile.
}
\label{fig:age}
\end{minipage}
\hfill
\begin{minipage}[t]{0.49\textwidth}
\centering
\includegraphics[scale=0.93]{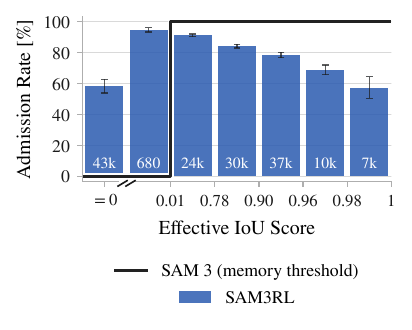}
\vspace{-1.5em}
\caption{\textbf{Admission rate of incoming frames} by SAM3RL on MOSEv2 against the frame's predicted IoU score scaled by $\max(2\sigma(o_t)-1,\,0)$.
SAM~3 stores only frames above 0.01.
Decisions are binned, counts in white, 95\% CIs.
}
\label{fig:admission}
\end{minipage}
\end{figure*}
\fi

\section{Discussion}

We presented SAM3RL, which learns memory control over a fixed-capacity bank with RL while keeping the backbone frozen. SAM3RL improves over SAM~3 on 11 of 12 benchmarks and ranks first on 10, with the largest gains on VOT(S), at virtually no runtime cost.
It is also the only method that consistently improves on video object segmentation.
Applying the same recipe to SAM~2 yields SAM2RL, which improves over SAM~2 on every benchmark and leads on 9.
These results suggest that memory control should be learned rather than designed, as learned control improves even the strongest trackers and, unlike fixed rules, remains effective across model generations.

\noindent\textbf{Limitations.}
First, the policy is tied to the 5\,FPS decision rate seen in training.
Deciding on every frame of 30\,FPS benchmarks still improves over SAM~3, but the average gain drops from 2.9 to 1.8 points.
We therefore match the training rate with the stride $s$, which requires the frame rate to be known in advance and keeps the skipped frames out of the memory bank.
The former is rarely restrictive in practice, as the frame rate is usually available.
Training at the native rate would remove this restriction, but most decisions would then concern near-identical frames.
Such a policy could nonetheless exploit the skipped frames and improve further on high-frame-rate benchmarks.
Second, the policy controls only the seven frames in the memory bank, while the nine object pointers are filled by a fixed rule.
Learning both jointly is a natural direction for future work.

\ifpreprint
\subsection*{Acknowledgments}

T. \v{C}\'{i}\v{z}ek, M. Straka, M. Kripner, and M. Schmid were supported by grant no.\ 25-18031S of the Czech Science Foundation (GA\v{C}R).
M. Schmid was also supported by Charles University project UNCE 24/SCI/008.
K. Janouskova was supported by the Grant Agency of the Czech Technical University in Prague, grant no. SGS26/074/OHK3/1T/13.
The authors acknowledge the support of the OP VVV funded project 
CZ.02.1.01/0.0/0.0/16\_019/0000765 ``Research Center for Informatics''.
The authors also thank EquiLibre Technologies, Inc.\ for supporting this research.
\fi

\bibliography{iclr2027_conference}
\bibliographystyle{iclr2027_conference}

\appendix
\clearpage
\crefalias{section}{appendix}
\crefalias{subsection}{appendix}

\section{Extended Results}
\label{suppl-results}

We extend the main model comparison from \Cref{tab:vot_datasets,tab:vot_competitions,tab:vos_results}. \Cref{tab:vot_competitions_sam2sam3,tab:vot_datasets_sam2sam3,tab:vos_results_sam2sam3} add three more benchmarks (GOT-10k, DAVIS$_{\text{2017}}$, and YouTube-VOS), together with the throughput of each method on VOT and VOS. We also report the compared methods in their original SAM~2 implementations alongside SAM2RL, our policy trained on the SAM~2 backbone. We discuss SAM~2 results in 
\Cref{sam2rl}, the three additional benchmarks in \Cref{add-bench}, and the VOS throughput in \Cref{vos-fps}.

\subsection{SAM2RL Performance}
\label{sam2rl}

In addition to the SAM 3 backbone results of \Cref{tab:vot_datasets,tab:vot_competitions,tab:vos_results}, we report the performance of compared methods in their original SAM~2 implementations. The results are presented alongside those of SAM2RL in \Cref{tab:vot_datasets_sam2sam3,tab:vot_competitions_sam2sam3,tab:vos_results_sam2sam3}, where SAM2RL 
is trained in the same way as SAM3RL. The MDP, the controller architecture, the LaSOT training corpus, the PPO hyperparameters, and the inference stride are unchanged, and only the frozen backbone differs. Unlike \Cref{sec:results}, this comparison includes GOT-10k, DAVIS$_{\text{2017}}$ and YouTube-VOS. The SAM~2 backbone leaves room for memory control to make a difference, and the compared methods still separate, which does not hold on SAM~3; see \Cref{add-bench} for further discussion.

On the VOT(S) challenges and VOT datasets, SAM2RL improves over SAM~2 on all nine benchmarks and ranks first on five. The gains are largest on the challenges, where it is the best method on four of five and second only to SAMITE on VOT-LT$_{\text{2021}}$. On DiDi it is ahead of every hand-crafted rule, while on LaSOT and LaSOT$_{\text{ext}}$ it trails SAMITE by less than one point. GOT-10k is the only tracking benchmark where the gain is negligible. SAM2RL also runs faster than SAM~2 itself and than every compared method, for the same reason as on SAM 3; see \Cref{sec:results}. On the segmentation benchmarks, SAM2RL ranks first on all six. It is the only method that improves on DAVIS$_{\text{2017}}$. On YT-VOS$_{\text{2019}}$, it matches the baseline, as does SAM2Long, while every other rule falls below it. Unlike on SAM 3, most hand-crafted rules do improve SAM 2 on SA-V, which supports the observation that they were designed to handle the specific weaknesses of the older backbone.

\begin{table*}[htb]
\setlength{\tabcolsep}{0.5em}
\caption{\textbf{Extended performance on VOT(S) challenges.} We report standard metrics for evaluated challenges. Deltas indicate improvement or decline relative to SAM 2 and SAM 3. \textbf{Bold} = best.}
\label{tab:vot_competitions_sam2sam3}
\begin{center}
\begin{tabular}{l c c c c c c}
\toprule
& & \multicolumn{3}{c}{Quality} & \multicolumn{2}{c}{F-score} \\
\cmidrule(lr){3-5} \cmidrule(lr){6-7}
Model & FPS & VOTS$_{\text{2026}}$ & VOTS$_{\text{2025}}$ & VOTS-RT$_{\text{2025}}$ & VOT-LT$_{\text{2021}}$ & VOT-LT$_{\text{2022}}$ \\
\midrule
SAM 3    & 15.3  & 66.4 & 73.4 & 69.1 & 77.1 & 62.8 \\
SAMURAI3  & 15.1 & 64.0\drop{2.4} & 72.2\drop{1.2} & 71.5\gain{2.4} & 77.4\gain{0.3} & 63.4\gain{0.6} \\
SAM3Long  & 9.3 & 67.9\gain{1.5} & 75.2\gain{1.8} & -- & 77.1\same & 64.5\gain{1.7} \\
DAM4SAM3   & 14.2& 66.7\gain{0.3} & 73.3\drop{0.1} & 69.5\gain{0.4} & 77.6\gain{0.5} & 63.4\gain{0.6} \\
SAMITE3    & 12.7& 65.4\drop{1.0} & 73.6\gain{0.2} & 72.7\gain{3.6} & 78.3\gain{1.2} & 66.2\gain{3.4} \\
HiM3SAM    & 11.8& 63.0\drop{3.4} & 72.5\drop{0.9} & 58.8\drop{10.3} & 77.1\same & 62.3\drop{0.5} \\
\textbf{SAM3RL (ours)} & \textbf{16.0}& \textbf{70.0}\gain{3.6} & \textbf{76.9}\gain{3.5} & \textbf{75.1}\gain{6.0} & \textbf{79.0}\gain{1.9} & \textbf{68.3}\gain{5.5} \\
\midrule
SAM 2.1 L  & 25.1 & 57.9 & 65.6 & 65.6 & 66.1 & 55.1 \\
SAMURAI    & 21.3 & 59.9\gain{2.0} & 68.2\gain{2.6} & 68.1\gain{2.5} & 75.8\gain{9.7} & 61.2\gain{6.1} \\
SAM2Long   & 11.0 & 61.0\gain{3.1} & 68.1\gain{2.5} & -- & 72.9\gain{6.8} & 57.5\gain{2.4} \\
DAM4SAM    & 16.7 & 61.7\gain{3.8} & 70.4\gain{4.8} & 70.5\gain{4.9} & 74.5\gain{8.4} & 60.9\gain{5.8} \\
SAMITE     & 16.3 & 61.1\gain{3.2} & 69.2\gain{3.6} & 67.1\gain{1.5} & \textbf{78.5}\gain{12.4} & 63.7\gain{8.6} \\
HiM2SAM    & 16.1 & 58.6\gain{0.7} & 67.8\gain{2.2} & 60.5\drop{5.1} & 73.9\gain{7.8} & 62.9\gain{7.8} \\
\textbf{SAM2RL (ours)} & \textbf{25.5}& \textbf{63.3}\gain{5.4} & \textbf{71.4}\gain{5.8} & \textbf{71.3}\gain{5.7} & 76.6\gain{10.5} & \textbf{64.6}\gain{9.5} \\
\bottomrule
\end{tabular}
\end{center}
\end{table*}

\begin{table*}[htb]
\setlength{\tabcolsep}{0.5em}
\caption{\textbf{Extended performance on VOT datasets.}  We report standard metrics for the evaluated datasets. Deltas indicate improvement or decline relative to SAM 2 and SAM 3. \textbf{Bold} = best.}
\label{tab:vot_datasets_sam2sam3}
\begin{center}
\begin{tabular}{l c c c c}
\toprule
& Quality & \multicolumn{2}{c}{AUC} & AO \\
\cmidrule(lr){2-2} \cmidrule(lr){3-4} \cmidrule(lr){5-5}
Model & DiDi & LaSOT & LaSOT$_{\text{ext}}$ & GOT-10k \\
\midrule
SAM 3   & 71.4 & 75.9 & 61.7 & 84.7 \\
SAMURAI3   & 71.8\gain{0.4} & 76.3\gain{0.4} & 63.5\gain{1.8} & \textbf{85.0}\gain{0.3} \\
SAM3Long   & 71.7\gain{0.3} & 76.1\gain{0.2} & 62.0\gain{0.3} & 84.9\gain{0.2} \\
DAM4SAM3   & 71.8\gain{0.4} & 77.3\gain{1.4} & \textbf{65.0}\gain{3.3} & 84.8\gain{0.1} \\
SAMITE3    & 71.8\gain{0.4} & 77.6\gain{1.7} & 64.7\gain{3.0} & 84.5\drop{0.2} \\
HiM3SAM    & 70.8\drop{0.6} & 75.6\drop{0.3} & 61.9\gain{0.2} & 84.3\drop{0.4} \\
\textbf{SAM3RL (ours) } & \textbf{72.0}\gain{0.6} & \textbf{78.4}\gain{2.5} & 64.7\gain{3.0} & 84.8\gain{0.1} \\
\midrule
SAM 2.1 L  & 64.8 & 69.9 & 56.9 & 80.6 \\
SAMURAI  & 67.8\gain{3.0} & 74.8\gain{4.9} & 60.0\gain{3.1} & \textbf{81.9}\gain{1.3} \\
SAM2Long & 66.2\gain{1.4} & 73.4\gain{3.5} & 59.1\gain{2.2} & 81.0\gain{0.4} \\
DAM4SAM  & 68.5\gain{3.7} & 75.5\gain{5.6} & 61.7\gain{4.8} & 81.3\gain{0.7} \\
SAMITE   & 68.6\gain{3.8} & \textbf{75.8}\gain{5.9} & \textbf{61.9}\gain{5.0} & 80.1\drop{0.5} \\
HiM2SAM  & 67.3\gain{2.5} & 74.3\gain{4.4} & 60.5\gain{3.6} & 75.8\drop{4.8} \\
\textbf{SAM2RL (ours)} & \textbf{69.1}\gain{4.3} & 75.4\gain{5.5} & 61.1\gain{4.2} & 80.7\gain{0.1} \\
\bottomrule
\end{tabular}
\end{center}
\end{table*}

\begin{table*}[t]
\setlength{\tabcolsep}{0.4em}
\caption{\textbf{Extended performance on VOS datasets.} We report standard metrics for the evaluated datasets. Deltas indicate improvement or decline relative to SAM 2 and SAM 3. \textbf{Bold} = best.
}
\label{tab:vos_results_sam2sam3}
\begin{center}
\begin{tabular}{l c c c c c c c }
\toprule
& & \multicolumn{1}{c}{$\mathcal{J\&\dot{F}}$} & \multicolumn{4}{c}{$\mathcal{J\&F}$} & {$\mathcal{G}$} \\
\cmidrule(lr){3-3} \cmidrule(lr){4-7} \cmidrule(lr){8-8}
Model & FPS & MOSE$_{\text{v}2}$ & SA-V$_{\text{val}}$ & SA-V$_{\text{test}}$ & LVOS$_{\text{v2}}$ & DAVIS$_{\text{2017}}$ & YT-VOS$_{\text{2019}}$  \\
\midrule
SAM 3   & \textbf{15.7} & 59.8& 82.9& 84.6& 88.7& \textbf{91.6}& \textbf{89.6} \\
SAMURAI3  & 15.2 & 56.6\drop{3.2}& 78.7\drop{4.2}& 81.4\drop{3.2}& 88.4\drop{0.3} & 91.3\drop{0.3}& 89.2\drop{0.4} \\
SAM3Long  & 9.4 & 59.3\drop{0.5}& 82.5\drop{0.4}& \textbf{85.0}\gain{0.4}& 89.0\gain{0.3} & 91.3\drop{0.3}& 89.5\drop{0.1} \\
DAM4SAM3  & 14.5 & 58.0\drop{1.8}& 80.4\drop{2.5}& 83.1\drop{1.5}& 88.9\gain{0.2}& \textbf{91.6}\same& 89.4\drop{0.2} \\
SAMITE3    & 12.8 & 53.4\drop{6.4}& 81.1\drop{1.8}& 82.2\drop{2.4}& 89.3\gain{0.6} & 90.7\drop{0.9} & 88.9\drop{0.7} \\
HiM3SAM    & 10.6 & 50.0\drop{9.8}& 81.1\drop{1.8}& 80.9\drop{3.7}& 86.6\drop{2.1} & 89.9\drop{1.7}& 88.5\drop{1.1} \\
\textbf{SAM3RL (ours)} & 15.4 & \textbf{60.9}\gain{1.1} & \textbf{85.4}\gain{2.5} & 84.6\same & \textbf{89.5\gain{0.8}} & 91.5\drop{0.1} & 89.5\drop{0.1} \\  
\midrule
SAM 2.1 L       & \textbf{25.0} & 49.4& 78.4& 79.8& 84.2& 89.8& \textbf{88.5} \\
SAMURAI    & 21.1 & 48.1\drop{1.3}& 79.3\gain{0.9}& 79.5\drop{0.3}& 86.3\gain{2.1}& 89.6\drop{0.2}& 88.1\drop{0.4} \\
SAM2Long   & 12.1& 50.7\gain{1.3}& 81.4\gain{3.0}& 80.7\gain{0.9}& 85.0\gain{0.8}& 89.8\same& \textbf{88.5}\same\\
DAM4SAM    & 17.9 & 50.5\gain{1.1}& 81.1\gain{2.7}& 81.9\gain{2.1}& 86.8\gain{2.6}& 88.4\drop{1.4}& 88.3\drop{0.2} \\
SAMITE    & 17.0& 47.0\drop{2.4}& 79.9\gain{1.5}& 80.1\gain{0.3}& 86.5\gain{2.3} & 88.2\drop{1.6}& 87.9\drop{0.6} \\
HiM2SAM    & 13.9& 42.7\drop{6.7} & 80.6\gain{2.2} & 80.2\gain{0.4}& 82.3\drop{1.9}& 87.1\drop{2.7}& 87.4\drop{1.1} \\
\textbf{SAM2RL (ours)} & 24.1 & \textbf{52.4}\gain{3.0} & \textbf{82.5}\gain{4.1} & \textbf{82.8}\gain{3.0} &  \textbf{86.9}\gain{2.7} & \textbf{90.7}\gain{0.9} & \textbf{88.5}\same \\
\bottomrule
\end{tabular}
\end{center}
\end{table*}

\subsection{Additional Benchmarks}
\label{add-bench}

We additionally report three benchmarks that are not part of the main comparison, GOT-10k, DAVIS$_{\text{2017}}$ and YouTube-VOS, in \Cref{tab:vot_datasets_sam2sam3,tab:vos_results_sam2sam3}. Performance is scored by the average overlap (AO) on GOT-10k, and by the overall score $\mathcal{G}$ on YouTube-VOS, which is the mean of $\mathcal{J}\&\mathcal{F}$ over categories seen and unseen during training. 

\textbf{GOT-10k}~\citep{8922619} is the shortest of all tracking benchmarks we evaluate on, with
an average sequence length of $127$ frames. In \Cref{tab:vot_datasets_sam2sam3}, five of the seven SAM~3 methods lie within $0.3$ of
each other, and the SAM~3 baseline is already at $84.7$. SAM3RL reaches $84.8$, $0.2$ behind SAMURAI3. This further highlights the previous findings that GOT-10k difficulty is shrinking for modern trackers~\citep{dam4sam_ijcv}.

\textbf{DAVIS$_\text{2017}$}~\citep{ponttuset20182017davischallengevideo,7780454} and \textbf{YouTube-VOS}~\citep{xu2018youtubevossequencetosequencevideoobject} are the
classic VOS benchmarks, on which modern methods are near
$90$~$\mathcal{J}\&\mathcal{F}$ and which contain mostly salient, dominant, and
isolated objects~\citep{ding2025mosev2challengingdatasetvideo}. In \Cref{tab:vos_results_sam2sam3}, the SAM~3 baseline already performs well on these datasets, leaving little room for memory-control methods to
improve it. Therefore, all methods remain within a few points of each
other, with the main exception of HiM3SAM. The small differences on these
benchmarks contrast with the larger gaps on other VOS datasets,
indicating that conventional short VOS
benchmarks do not provide a discriminative test of memory management. 

Note that the benchmark counts in the main paper, such as ``first on 10 of 12'', are over the twelve benchmarks of \Cref{tab:vot_datasets,tab:vot_competitions,tab:vos_results} and exclude the three discussed here, because on SAM~3, none of the three benchmarks separates the compared methods. We nonetheless keep them in \Cref{tab:vot_datasets_sam2sam3,tab:vos_results_sam2sam3}, since on the weaker SAM~2 backbone they still separate the methods, and \Cref{sam2rl} evaluates SAM2RL performance over all fifteen.

\subsection{VOS Throughput}
\label{vos-fps}

In addition to the VOTS throughput reported in \Cref{sec:results}, we measure VOS throughput on MOSE$_{\text{v2}}$ and list it in \Cref{tab:vos_results_sam2sam3}. The protocol is the same as on VOTS, but 
MOSE$_{\text{v2}}$ videos are annotated at an average of 5\,FPS, so both SAM2RL and SAM3RL update their memory on every frame with $s = 1$, unlike every-6-frames updates on VOTS with $s = 6$. SAM3RL falls slightly behind SAM~3 at $15.4$ vs $15.7$ FPS, but remains faster than all other methods. The same holds for SAM2RL, which runs at $24.1$ against $25.0$\,FPS for SAM 2.1 L, and is the fastest after the baseline.

\clearpage
\section{Tracker Evaluation Details and Reproducibility}
\label{suppl-eval}

\noindent\textbf{Benchmarks and metrics.}
\Cref{tab:datasets} lists every dataset we evaluate on, with the number of sequences, objects, frame rate, and length. Performance is measured by the tracking quality~\citep{10350713} on the VOTS challenges and on DiDi, and by the F-score~\citep{9054960} on the VOT-LT challenges. On LaSOT and LaSOT$_{\text{ext}}$, it is measured by the area under the success rate curve (AUC)~\citep{7001050}. On VOS benchmarks, it is measured by $\mathcal{J}\&\mathcal{F}$~\citep{ponttuset20182017davischallengevideo}, the mean of the region similarity $\mathcal{J}$ and the contour accuracy $\mathcal{F}$. MOSE$_{\text{v2}}$ replaces $\mathcal{F}$ by an adaptive boundary accuracy $\dot{\mathcal{F}}$ that rescales the tolerance with object size.

We score LaSOT and LaSOT$_{\text{ext}}$ with the PyTracking evaluation code~\citep{pytracking} (commit \texttt{9cba77e}), which implements the AUC protocol of the official LaSOT toolkit~\citep{fan2019lasothighqualitybenchmarklargescale}. Our SAM~2.1 runs reproduce the published AUC of 70.0 on LaSOT and 56.9 on LaSOT$_{\text{ext}}$~\citep{ding2025sam2longenhancingsam2,dam4sam_ijcv} to within 0.1. Some evaluators implement the protocol differently (\eg the got10k toolkit~\citep{8922619} and the pysot toolkit~\citep{pysot}), as does the evaluation code in the SAMURAI codebase~\citep{yang2024samuraiadaptingsegmentmodel}. For instance, on LaSOT$_{\text{ext}}$, their implementations inflate scores of every method we report by up to 1.7 AUC compared to the official protocol. GOT-10k, MOSE$_{\text{v2}}$, and all VOTS challenges are scored on their evaluation servers. DiDi and VOT-LT are scored with the VOT toolkit~\citep{CEHOVINZAJC2020100623}.

\begin{table*}[htb]
\caption{\textbf{Evaluation benchmarks} with VOT(S) challenges (top), VOT datasets (middle), VOS datasets (bottom). Seq. and Obj. are the numbers of
sequences and annotated targets, respectively. FPS is the annotation frame rate, Frames is the
average number of frames per sequence, and $s$ is the stride defined in \Cref{sec:learning}. The three length columns give the average,
shortest, and longest sequence duration in seconds. SA-V videos are recorded at 24 FPS, but annotated at 6.}
\label{tab:datasets}
\begin{center}
\setlength{\tabcolsep}{0.34em}
\begin{tabular}{l r r c r r r r r}
\toprule
& & & & & & \multicolumn{3}{c}{Length (s)} \\
\cmidrule(lr){7-9}
Dataset & Seq. & Obj. & FPS & $s$ & Frames & Avg. & Min & Max \\
\midrule
VOTS$_{\text{2026}}$~\citep{kristan2026the}  & 170 & 375 & 30 & 6 & 1860 & 62.0 & 2.1 & 356.6 \\
VOTS(-RT)$_{\text{2025}}$~\citep{10350713,11375669}  & 144 & 341 & 30 & 6 & 2074 & 69.1 & 2.1 & 356.6 \\
VOT-LT$_{\text{2021}}$~\citep{9607833}  & 50 & 50 & 30 & 6 & 4306 & 143.5 & 46.3 & 990.0 \\
VOT-LT$_{\text{2022}}$~\citep{10.1007/978-3-031-25085-9_25}  & 50 & 50 & 30 & 6 & 3366 & 112.2 & 32.6 & 356.6 \\
\midrule
DiDi~\citep{dam4sam_ijcv}  & 180 & 180 & 30 & 6 & 1527 & 50.9 & 2.1 & 333.3 \\ 
LaSOT~\citep{fan2019lasothighqualitybenchmarklargescale}  & 280 & 280 & 30 & 6 & 2448 & 81.6 & 33.3 & 333.3 \\
LaSOT$_{\text{ext}}$~\citep{fan2020lasothighqualitylargescalesingle}  & 150 & 150 & 30 & 6 & 2394 & 79.8 & 66.8 & 90.1 \\
GOT-10k~\citep{8922619} & 180 & 180 & 10 & 2 & 127 & 12.7 & 5.1 & 92.0 \\
\midrule
MOSE$_{\text{v2}}$~\citep{ding2025mosev2challengingdatasetvideo}  & 433 & 575 & 5 & 1 & 154 & 30.7 & 4.8 & 488.0 \\
SA-V$_{\text{val}}$~\citep{ravi2024sam2segmentimages}  & 155 & 293 & 24\,/\,6 & 5 & 414 & 17.2 & 2.2 & 38.4 \\
SA-V$_{\text{test}}$~\citep{ravi2024sam2segmentimages}  & 150 & 278 & 24\,/\,6 & 5 & 426 & 17.7 & 2.7 & 36.1 \\
LVOS$_{\text{v2}}$~\citep{hong2024lvosbenchmarklargescalelongterm}  & 140 & 239 & 6 & 1 & 472 & 78.6 & 19.2 & 356.7 \\
DAVIS$_{\text{2017}}$~\citep{ponttuset20182017davischallengevideo}  & 30 & 61 & 24 & 5 & 67 & 2.8 & 1.4 & 4.3 \\
YouTube-VOS$_{\text{2019}}$~\citep{xu2018youtubevossequencetosequencevideoobject}   & 507 & 1063 & 6 & 1 & 27 & 4.5 & 0.7 & 6.0 \\
\bottomrule
\end{tabular}
\end{center}
\end{table*}

\noindent\textbf{Previously published results} for the compared methods and baselines are listed in \Cref{tab:cross-paper2} and \Cref{tab:cross_paper}, next to our own measurements. Most agree
within noise, while some differ more. We attribute these differences to
evaluation protocols that vary between papers (\eg mentioned LaSOT$_{\text{ext}}$ case), to unspecified prompt masks, and
to differences in hardware and libraries. Interestingly, even SAM~2 baseline numbers are not consistent between previous work. 

\begin{table*}[t]
\setlength{\tabcolsep}{0.5em}
\caption{\textbf{Performance on tracking benchmarks across previous work.} For each work, we list the numbers for baselines as reported there (whether independently measured, or cited from prior work). The last block reports our own measurements. Hyphen means the benchmark was not reported.}
\label{tab:cross-paper2}
\begin{center}
\begin{tabular}{c l c c c c c}
\toprule
& & \multicolumn{2}{c}{Quality} & \multicolumn{2}{c}{AUC} & AO  \\
\cmidrule(lr){3-4} \cmidrule(lr){5-6} \cmidrule(lr){7-7} 
Source & Model & DiDi & VOTS$_{\text{2025}}$ & LaSOT & LaSOT$_{\text{ext}}$ & GOT-10k \\
\midrule
\multirow{2}{*}{\rotatebox{0}{\makecell{SAMURAI\\\citep{yang2024samuraiadaptingsegmentmodel}}}}
 & SAM 2.1 L  &-- &-- &68.5 &58.6 &--  \\
 & SAMURAI    &-- &-- &74.2 &61.0 &81.7 \\
\midrule
\multirow{2}{*}{\rotatebox{0}{\makecell{SAM2Long\\\citep{ding2025sam2longenhancingsam2}}}}
 & SAM 2.1 L &-- &-- & 70.0& 56.9& 80.7 \\
 & SAM2Long  &-- &-- & 73.9 & 60.9 & 81.1 \\
\midrule
 \multirow{3}{*}{\rotatebox{0}{\makecell{SAMITE\\\citep{xu2025samitepositionpromptedsam2}}}}
 & SAM 2.1 L &-- &-- & 68.5& 58.6  & -- \\
 & SAMURAI   &-- &-- & 74.2& 61.0  & -- \\
 & SAMITE    &-- &-- & 74.7 & 62.1 & -- \\
\midrule
 \multirow{5}{*}{\rotatebox{0}{\makecell{HiM2SAM\\\citep{chen2025him2samenhancingsam2hierarchical}}}}
 & SAM 2.1 L &-- &-- & 68.5& 58.6& -- \\
   & SAMURAI &-- &-- & 74.2& 61.0& -- \\
   & DAM4SAM &-- &-- & 75.1& 60.9& -- \\
     & SAM2Long &-- &-- & 73.9& 60.9& -- \\
 & HiM2SAM &-- &-- & 75.1 & 62.8 & -- \\
\midrule
\multirow{4}{*}{\rotatebox{0}{\makecell{DAM4SAM\\\citep{dam4sam_ijcv}}}}
 & SAM 2.1 L & 64.9& 66.1 & 70.0& 56.9&80.7 \\
 & SAMURAI & 68.0 &-- & --& --&-- \\
 & DAM4SAM & 69.4& 71.1 & 75.1 & 60.9 & 81.1\\
 & SAM2Long & 64.6 &-- & --& --&-- \\
\midrule
\multirow{7}{*}{\rotatebox{0}{\textbf{Ours}}}
&SAM 2.1 L  & 64.8 & 65.6 & 69.9 & 56.9 & 80.6 \\
&SAMURAI  &  67.8& 68.2 &  74.8&  60.0& 81.9 \\
&DAM4SAM & 68.5 & 70.4  & 75.5 & 61.7 &81.3 \\
&SAM2Long &66.2 & 68.1 & 73.4 &59.1 &81.0 \\
&SAMITE & 68.6 & 69.2 & 75.8 & 61.9 & 80.1 \\
&HiM2SAM & 67.3 & 67.8 & 74.3 & 60.5 & 75.8 \\
&SAM 3  & 71.4& 73.4  & 75.9 & 61.7 &84.7 \\
\bottomrule
\end{tabular}
\end{center}
\end{table*}
\begin{table*}[t]
\small
\setlength{\tabcolsep}{0.29em}
\caption{\textbf{Performance on VOS datasets across previous work.}  For each work, we list the numbers for baselines as reported there (whether independently measured, or cited from prior work). The last block presents our own measurements. Hyphen means the benchmark was not reported.}
\label{tab:cross_paper}
\begin{center}
\begin{tabular}{c l c c c c c c c}
\toprule
& & \multicolumn{1}{c}{$\mathcal{J\&\dot{F}}$} & \multicolumn{4}{c}{$\mathcal{J\&F}$} & {$\mathcal{G}$} \\
\cmidrule(lr){3-3} \cmidrule(lr){4-7} \cmidrule(lr){8-8} 
Source& Model & MOSE$_{\text{v2}}$ & SA-V$_{\text{val}}$ & SA-V$_{\text{test}}$ & LVOS$_{\text{v2}}$ & DAVIS$_{\text{2017}}$ & YT-VOS$_{\text{2019}}$  \\
\midrule

\multirow{4}{*}{\small\rotatebox{0}{\makecell{MOSE$_{\text{v}2}$\\\citep{ding2025mosev2challengingdatasetvideo}}}}
 & SAM 2.1 L        & 49.4 & -- & 78.4 & 84.0 & 90.7 & 89.3  \\
 & SAMURAI     & 51.1 & -- & -- & -- & -- & --  \\
 & DAM4SAM     & 51.2 & -- & -- & -- & -- & --  \\
 & SAM2Long    & 51.5 & -- & 81.2 & 85.3 & 88.8 & 90.2  \\
\midrule

\multirow{2}{*}{\small\rotatebox{0}{\makecell{DAM4SAM\\\citep{dam4sam_ijcv}}}}
 & SAM 2.1 L      & --& --& --& 80.6& --& --&  \\
 & DAM4SAM        & --& --& --& 85.5& --& --&  \\
\midrule

\multirow{2}{*}{\small\rotatebox{0}{\makecell{SAM2Long\\\citep{ding2025sam2longenhancingsam2}} }}
 & SAM 2.1 L       & -- & 78.6& 79.6& 84.0& --& -- \\
 & SAM2Long        & -- & 81.1& 81.2& 85.3& --& -- \\
\midrule

\multirow{4}{*}{\small\rotatebox{0}{\makecell{SAM3\\\citep{carion2026sam3segmentconcepts}}}}
 & SAM 2.1 L      & 47.9 & 77.9 & 78.4 & 79.6 & 90.7 & --  \\
 & SAMURAI        & 51.1 & 79.8 & 80.0 & 84.2 & 89.9 & --  \\
 & SAM2Long       & 51.5 & 81.2 & 81.1 & 85.9 & 91.4 & --  \\
 & SAM3           & 60.3 & 83.5 & 84.4 & 88.5 & 92.2 & --  \\
\midrule

\multirow{5}{*}{\small\rotatebox{0}{\makecell{\textbf{Ours}}}}
 & SAM 2.1 L       & 49.4& 78.4& 79.8& 84.2& 89.8& 88.5 \\
 & SAMURAI     & 48.1& 79.3& 79.5& 86.3& 89.6& 88.1 \\
 & DAM4SAM    & 50.5& 81.1& 81.9 & 86.8 & 88.4 & 88.3 \\
 & SAM2Long   & 50.7& 81.4& 80.7& 85.0& 89.8& 88.5 \\
 & SAM3    &59.8& 82.9& 84.6& 88.7& 91.6& 89.6 \\

\bottomrule
\end{tabular}
\end{center}
\end{table*}

\noindent\textbf{Initialization prompts} used for the trackers in our evaluation are as follows. We use segmentation masks wherever such are available. A mask localizes the target more precisely than a box. SAM~2 reports $79.3$ against $74.4$ $\mathcal{J\&F}$ averaged over 17 video datasets when the first frame is prompted by a ground-truth mask rather than a box \citep{ravi2024sam2segmentimages}. SAM~3 then inherits masks as main prompts in its evaluation~\citep{carion2026sam3segmentconcepts}.

We do not use mask prompting only where a method or a challenge protocol prescribes otherwise.
On VOS benchmarks, every method is initialized with the ground-truth mask of the dataset. On LaSOT, LaSOT$_\text{ext}$, and GOT-10k, we use the masks from the set released by~\citet{dam4sam_ijcv}, so that the prompt is reproducible from a public file. The single exception for these VOT datasets is SAMURAI(3), which we prompt with the box, as its authors do and as its published numbers were obtained.\footnote{We also ran SAMURAI with the mask prompt on LaSOT. The box arm is better by
$0.3$ AUC.} DiDi ships its own first-frame masks, and we use them. On VOT-LT$_{\text{2021}}$ and VOT-LT$_{\text{2022}}$, we prompt with the ground-truth box. That is the official protocol of both challenges, and no public mask set exists either. On VOTS the toolkit delivers the ground-truth mask of each target.

\paragraph{Object handling.}
We evaluate trackers in a per-object rather than joint setup. Each object is
initialized with its first non-empty annotation and propagated forward from that
frame. The tracker state is reset between objects. This is the only mode all
methods support, since SAMURAI, DAM4SAM, SAM2Long, SAMITE, and HiM2SAM are
single-object trackers by construction. It is also the better mode for SAM~3,
which gains up to $2.8$ $\mathcal{J}\&\mathcal{F}$ over joint tracking on SA-V$_{\text{test}}$. VOTS multi-target sequences are run in the sequential single-object mode
of the toolkit, one tracker process per target.

\paragraph{Port implementation.}
None of the compared methods has an official SAM~3 implementation, so we port each to that backbone. We replace the SAM~2 video predictor with the SAM~3 tracker and leave the method's code, hyperparameters, and memory layout unchanged. Additionally, we implement the striding described in \Cref{sec:learning} for DAM4SAM~\citep{dam4sam_ijcv}. This follows its VOS protocol, which \citet{dam4sam_ijcv} describe in the paper.
The official code for this feature was never released, so we reproduce it from the description.
\ifpreprint
We will release the port implementations together with our code.
\fi

\paragraph{SAM2Long and SAM3Long are not online trackers.}
\citet{ding2025sam2longenhancingsam2} keep several memory pathways and decodes the video only after tracking ends, along the pathway that leads in the last frame. We preserve this behavior for both SAM2Long and SAM3Long, since changing them to fulfill the causality would be a different algorithm. However, the comparison on the tracking benchmarks is uneven, where every other method, ours included, is online. As a result, SAM(2/3)Long are excluded from VOTS-RT$_{\text{2025}}$, where the answer for frame $t$ is due before frame $t+1$ arrives.

\paragraph{Hardware.}
All evaluations are computed on NVIDIA A100 40\,GB GPUs unless specified
otherwise. Throughput is also measured on the A100, as the mean over three runs. It times
only model inference, with all the data preprocessing and postprocessing being excluded.

Notably, some methods utilize much more RAM or GPU memory than others. For instance, the longest VOT-LT$_{\text{2021}}$ sequence, decoded to CPU, takes $348$\,GB for most methods, but SAM(2/3)Long need $600$\,GB on that sequence, since they store three pathways per frame. Similarly, the SAMITE prototype bank of up to $30$ frames exceeds the $40$\,GB of the A100 on MOSE$_{\text{v2}}$, so the SAMITE family is run on NVIDIA H200 NVL. Throughput is nonetheless measured on the A100 for SAMITE and SAM2Long, since the sequences used for timing fit into its memory.

\clearpage
\section{Race Condition in SAM~3}
\label{suppl-race-condition}

The SAM~3 video predictor offloads per-frame tracking state to pageable host memory
with \texttt{non\_blocking=True} and then reads that buffer on the CPU without
synchronizing, which PyTorch documents as undefined behavior. Within a single
\texttt{add\_new\_points\_or\_box} call, the asynchronous copy of the
conditioning-frame mask is launched and the same tensor is interpolated on the
host a few lines later. If the copy did not land, the host will read the
leftovers of the previous sequence in the process. The corrupted mask is encoded
into the memory bank as frame~0, and because long-term tracking never
re-initializes, the sequence is lost from frame~1 rather than degrading over
time. The failure requires box initialization, CPU state offloading, and a
preceding sequence in the same process. We fix it by forcing CUDA$\rightarrow$CPU copies to be synchronous.
 
The impact of the race condition was seen most severely on the VOT-LT$_{\text{2021}}$ challenge, where all methods are box-prompted, and on SAMURAI3, which is also box-prompted on all benchmarks; see \Cref{suppl-eval}. Table~\ref{tab:supp_d2h_lt2021} shows VOT-LT$_{2021}$ before and after the fix. Four
baselines gain between $5.1$ and $12.3$~F. DAM4SAM3 and SAMITE3 are
byte-identical on all fifty sequences before and after the fix. 
We report performance changes for SAMURAI3 in Table~\ref{tab:supp_d2h_samurai}. Post-fix, SAMURAI3 recovers
$+13.1$ AUC on LaSOT and $+7.7$ on LaSOT$_{\text{ext}}$. All other numbers in this work are post-fix. 

\begin{table*}[htb]
  \small
  \begin{minipage}[t]{0.47\textwidth}
    \caption{\textbf{Effect of forcing CUDA$\rightarrow$CPU copies to be synchronous on
    VOT-LT$_{\text{2021}}$.} DAM4SAM3 and SAMITE3 were unaffected.}
    \label{tab:supp_d2h_lt2021}
    \begin{center}
    \begin{tabular}{lccc}
      \toprule
      Method & pre-fix & after & $\Delta$ \\
      \midrule
      HiM3SAM   & 64.8 & 77.1 & \color{fgreen}{+12.3} \\
      SAM3Long  & 69.1 & 77.1 & \color{fgreen}{+8.0} \\
      SAM~3     & 71.4 & 77.1 & \color{fgreen}{+5.7} \\
      SAMURAI3  & 72.3 & 77.4 & \color{fgreen}{+5.1} \\
      DAM4SAM3  & 77.6 & 77.6 & \color{gray}{±0.0} \\
      SAMITE3   & 78.3 & 78.3 & \color{gray}{±0.0} \\
      \bottomrule
    \end{tabular}
    \end{center}
  \end{minipage}
  \hfill
  \begin{minipage}[t]{0.48\textwidth}
    \caption{\textbf{SAMURAI3 performance on the VOT datasets exposed to
    the race condition before and after the fix.} Unlike other trackers, SAMURAI3
    is always prompted by bounding boxes; see \Cref{suppl-eval}. Therefore, it was the only one affected outside
    VOT-LT$_{\text{2021}}$.}
    \label{tab:supp_d2h_samurai}
    
    \begin{center}
    \setlength{\tabcolsep}{0.4em}
    \begin{tabular}{llccc}
      \toprule
      Split & Metric & pre-fix & after & $\Delta$ \\
      \midrule
      GOT-10k          & AO & 82.8 & 85.0 & \color{fgreen}{+2.2} \\
      LaSOT          & AUC & 63.2 & 76.3 & \color{fgreen}{+13.1} \\
      LaSOT$_{\text{ext}}$  & AUC & 55.8 & 63.5 & \color{fgreen}{+7.7} \\
      VOT-LT$_{2021}$     & F   & 72.3 & 77.4 & \color{fgreen}{+5.1} \\
      \bottomrule
    \end{tabular}
    \end{center}
  \end{minipage}
\end{table*}

\clearpage
\section{Training Setup}
\label{suppl-hyperparameters}

We provide further details on the training setup in addition to those in \Cref{sec:method}. All training hyperparameters are listed in \Cref{tab:supp_hparams}.

\noindent\textbf{Pre-computed embeddings.}
The image encoder is frozen and its output does not depend on the policy.
Thus, we compute the image embeddings of the whole training corpus once and cache them on disk.
During rollouts, only the tracker components after the image encoder are run.

\noindent\textbf{Batch composition.}
Each rank runs 64 parallel environments, each tracking one object. They come from 16 videos, with each object copied four times.
We truncate the batch to the length of its shortest video, so that all environments are synchronized and process the same frame index at every step. To discard few frames, videos are grouped into batches by similar length, with each length jittered by up to 16 steps so that the same videos are not always grouped together.

\noindent\textbf{Truncation.}
Apart from this batch truncation, episodes are not capped, whereas a rollout collects 376 steps. An episode that crosses the rollout boundary continues in the next iteration, and at the cut its return is bootstrapped with the critic's value of the last state. The end of the truncated batch, by contrast, is terminal and is not bootstrapped.

\begin{table}[htb]
\caption{\textbf{Training hyperparameters for SAM2RL and SAM3RL.}
Both backbones are frozen. Only the actor--critic is optimized on pre-extracted image embeddings.}
\label{tab:supp_hparams}
\begin{center}
\setlength{\tabcolsep}{6pt}
\begin{tabular}{ll}
\toprule
Hyperparameter & Value \\
\midrule
\multicolumn{2}{l}{\emph{Task}} \\
\quad Backbones                       & SAM 3, frozen; image size $1008\times1008$, bf16 \\
\quad                                & SAM 2.1 L, frozen; image size $1024\times1024$, bf16 \\
\quad Memory bank size               & 7 slots \\
\quad Action space                   & 7 (evict one slot after appending the current frame) \\
\quad Trainable parameters           & 6.45\,M (actor 3.23\,M, critic 3.22\,M) \\
\addlinespace
\multicolumn{2}{l}{\emph{Data}} \\
\quad Corpus                         & LaSOT$_{\text{5fps}}$ (stride-6 subsample of LaSOT, 5\,fps) \\
\quad Split                          & 896 train / 224 validation sequences (of 1120) \\
\quad Episode length                 & full video, truncated to the shortest video in batch \\
\quad Batch composition              & similar-length batching, length jitter 16 \\
\quad Environments per rank          & 64, drawn from 16 unique videos (object multiplier 4) \\
\addlinespace
\multicolumn{2}{l}{\emph{PPO}} \\
\quad Iterations                     & 400 \\
\quad Parallel environments          & 256 ($64\times4$ ranks) \\
\quad Rollout length             & 376 steps ($96{,}256$ transitions per iteration) \\
\quad Update epochs                  & 4 \\
\quad Minibatch size                 & 1024 \\
\quad Optimizer                      & Adam, $\varepsilon=10^{-5}$, weight decay 0 \\
\quad Learning rate                  & $3\times10^{-4}$, constant (no annealing) \\
\quad Discount $\gamma$              & 1.0 \\
\quad Task horizon                   & finite (episode end is terminal) \\
\quad Advantage estimator            & GAE, $\lambda = 1$ (Monte-Carlo return minus critic baseline) \\
\quad Policy clip $\epsilon$         & 0.1 \\
\quad Target KL (early stop)         & 0.02 \\
\quad Entropy coefficient            & $10^{-4}$, constant (no annealing) \\
\quad Value loss                     & Huber, $\delta=5$; coefficient 0.5 \\
\quad Advantage normalization        & yes (per minibatch) \\
\quad Gradient clipping              & $\|g\|_2\leq2.0$, applied per head \\
\addlinespace
\multicolumn{2}{l}{\emph{Compute}} \\
\quad Hardware                       & 24 hours on $4\times$ H200 (one node, 32 CPU, 256\,GB), DDP \\
\quad Seeds                          & 1, 2, 3 (main results use seed 1, the best on local validation) \\
\bottomrule
\end{tabular}
\end{center}
\end{table}

\clearpage

\clearpage
\section{Ablations}
\label{suppl:ablations}

All ablations are based on the SAM3RL training setup described in \Cref{sec:method} and \Cref{suppl-hyperparameters}, and the evaluation protocols follow \Cref{sec:results}.
We report ablation results as the mean over three training seeds to reduce the impact of training variance.
In contrast, the main results in \Cref{tab:vot_competitions,tab:vot_datasets,tab:vos_results} use the single best of these seeds, selected on our validation split without using the test benchmarks, as they aim to show the best model we can obtain.
The effects of the actor input features, the object pointer selection rule, and the reward normalization are presented in \Cref{abl-actor}, \Cref{abl-objectpointers}, and \Cref{abl-reward}, respectively.

\subsection{Actor Input Features}
\label{abl-actor}

The state $s_t$ groups the per-slot features into four disjoint groups: temporal, confidence, semantic, and geometry; see~\Cref{sec:mdp}. 
We test whether each group is necessary or sufficient with delete-one and keep-one ablations, respectively. The former removes one group from the actor input, while the latter keeps exactly one.
Everything else, including the critic input, is unchanged.
The groups differ in the input size, with the baseline actor input having 1,560 dimensions after the pointer projection, of which semantic take 1,152, temporal 256, geometry 120, and confidence 32. 
The performance is presented in~\Cref{tab:delete_actor_input_ablation_mean,tab:keep_actor_input_ablation_mean} as a mean over three seeds for each arm. No single group is critically necessary, and no single group is sufficient either. All arms still improve over SAM~3, except the keep-one arms on MOSE$_{\text{v2}}$, yet none matches the all-inputs actor across all benchmarks. 

\begin{table*}[htb]
\setlength{\tabcolsep}{0.5em}
\caption{\textbf{Delete-one actor input feature group ablation.} Each row removes one feature group from the actor input.
Everything else, including the critic input, is unchanged.
Every entry is the mean over three seeds, with the standard error of the mean in subscript. Cell color encodes the difference to the full-input SAM3RL mean, {\color{gainC}green} for a gain and {\color{dropC}red} for a drop, darker for a larger difference. \textbf{Bold} indicates the best.}
\label{tab:delete_actor_input_ablation_mean}
\begin{center}
\begin{tabular}{l c c c c c}
\toprule
& Quality & F-score & \multicolumn{2}{c}{AUC} & $\mathcal{J}\&\mathcal{F}$ \\
\cmidrule(lr){2-2} \cmidrule(lr){3-3} \cmidrule(lr){4-5} \cmidrule(lr){6-6}
Actor input & VOTS$_{\text{2026}}$ & VOT-LT$_{\text{2022}}$ & LaSOT & LaSOT$_{\text{ext}}$ & MOSE$_{\text{v2}}$ \\
\midrule
\textbf{SAM3RL} (all inputs) & \rcell{\textbf{69.5}}{0.2} & \rcell{68.5}{0.3} & \rcell{\textbf{78.0}}{0.3} & \rcell{\textbf{64.6}}{0.1} & \rcell{60.6}{0.3} \\
\midrule
w/o temporal  & \dcell{69.1}{0.3}{-0.4} & \dcell{\textbf{68.9}}{0.5}{+0.4} & \dcell{\textbf{78.0}}{0.1}{0} & \dcell{64.3}{0.3}{-0.3} & \dcell{\textbf{60.7}}{0.3}{+0.1} \\
w/o confidence    & \dcell{69.2}{0.2}{-0.3} & \dcell{68.4}{0.4}{-0.1} & \dcell{77.9}{0.1}{-0.1} & \dcell{64.1}{0.1}{-0.5} & \dcell{60.4}{0.1}{-0.2} \\
w/o semantic & \dcell{69.1}{0.4}{-0.4} & \dcell{67.3}{0.4}{-1.2} & \dcell{\textbf{78.0}}{0.2}{0} & \dcell{64.5}{0.2}{-0.1} & \dcell{60.2}{0.2}{-0.4} \\
w/o geometry  & \dcell{69.2}{0.2}{-0.3} & \dcell{68.2}{0.2}{-0.3} & \dcell{77.8}{0.1}{-0.2} & \dcell{64.5}{0.2}{-0.1} & \dcell{59.7}{0.2}{-0.9} \\
\bottomrule
\end{tabular}
\end{center}
\end{table*}
\begin{table*}[htb]
\setlength{\tabcolsep}{0.5em}
\caption{\textbf{Keep-one actor input feature group ablation.} Each row keeps one feature group in the actor input and removes the other three.
Everything else, including the critic input, is unchanged.
Every entry is the mean over three seeds, with the standard error of the mean in subscript. Cell color encodes the difference to the full-input SAM3RL mean, {\color{gainC}green} for a gain and {\color{dropC}red} for a drop, darker for a larger difference. \textbf{Bold} indicates the best.}
\label{tab:keep_actor_input_ablation_mean}
\begin{center}
\begin{tabular}{l c c c c c}
\toprule
& Quality & F-score & \multicolumn{2}{c}{AUC} & $\mathcal{J}\&\mathcal{F}$ \\
\cmidrule(lr){2-2} \cmidrule(lr){3-3} \cmidrule(lr){4-5} \cmidrule(lr){6-6}
Actor input & VOTS$_{\text{2026}}$ & VOT-LT$_{\text{2022}}$ & LaSOT & LaSOT$_{\text{ext}}$ & MOSE$_{\text{v2}}$ \\
\midrule
\textbf{SAM3RL} (all inputs) & \rcell{\textbf{69.5}}{0.2} & \rcell{68.5}{0.3} & \rcell{78.0}{0.3} & \rcell{64.6}{0.1} & \rcell{\textbf{60.6}}{0.3} \\
\midrule
only temporal  & \dcell{68.7}{0.3}{-0.8} & \dcell{68.3}{0.0}{-0.2} & \dcell{77.6}{0.2}{-0.4} & \dcell{64.1}{0.1}{-0.5} & \dcell{59.2}{0.1}{-1.4} \\
only confidence    & \dcell{68.7}{0.2}{-0.8} & \dcell{66.3}{0.2}{-2.2} & \dcell{77.7}{0.2}{-0.3} & \dcell{\textbf{64.8}}{0.1}{+0.2} & \dcell{60.1}{0.6}{-0.5} \\
only semantic & \dcell{69.1}{0.4}{-0.4} & \dcell{\textbf{68.8}}{0.3}{+0.3} & \dcell{\textbf{78.1}}{0.3}{+0.1} & \dcell{63.7}{0.4}{-0.9} & \dcell{59.8}{0.2}{-0.8} \\
only geometry  & \dcell{68.8}{0.2}{-0.7} & \dcell{67.6}{0.6}{-0.9} & \dcell{77.7}{0.2}{-0.3} & \dcell{63.9}{0.2}{-0.7} & \dcell{58.9}{0.8}{-1.7} \\
\bottomrule
\end{tabular}
\end{center}
\end{table*}

\subsection{Object Pointer Selection}
\label{abl-objectpointers}
\begin{figure*}[htb]
\begin{minipage}[t]{0.49\textwidth}
\centering
\includegraphics[scale=0.98]{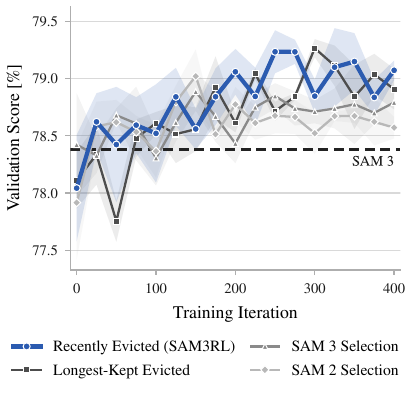}
\caption{\textbf{Object pointer selection.} Validation score on the LaSOT validation split during SAM3RL training; mean over three seeds with a $\pm$SEM band. The arms differ only in the rule that fills the nine object pointers outside the bank.}
\label{fig:ablation_pointers}
\end{minipage}%
\hfill
\begin{minipage}[t]{0.49\textwidth}
\centering
\includegraphics[scale=0.98]{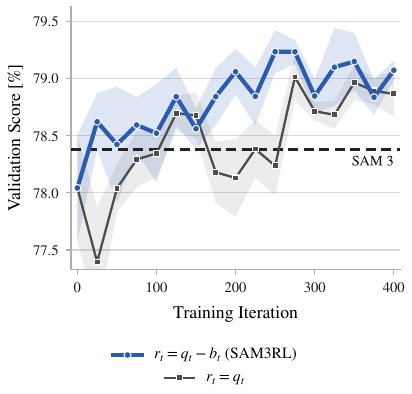}
\caption{\textbf{Reward normalization.} Validation score on the LaSOT validation split during training; mean over three seeds with a $\pm$SEM band. The arms differ only in the $b_t$ normalized reward, $r_t = q_t - b_t$ against the raw quality $r_t = q_t$.}
\label{fig:ablation_reward}
\end{minipage}
\end{figure*}

Besides the seven memory features, the memory attention of SAM~2 and SAM~3 also reads up to sixteen object pointers, one of which always comes from the conditioning frame; see \Cref{sec:method}.
SAM~2 takes the remaining pointers from the 15 most recent frames, whereas SAM~3 takes them from the most recent frames whose effective IoU score exceeds a threshold, always including the previous frame.
SAM3RL instead uses a \emph{recently evicted} rule, which reads the pointers of the conditioning frame and of the six frames currently in the memory bank, and fills the remaining nine slots with the frames most recently evicted from the bank.
We also consider a \emph{longest-kept} rule, which ranks evicted frames by the number of decisions they remained in the memory bank and selects the pointers of the top nine.
Unlike the rules of SAM~2 and SAM~3, both of our rules depend on the agent's actions.
\Cref{fig:ablation_pointers} compares all four rules within SAM3RL, reporting the LaSOT validation score of each rule every 25 training iterations; all other components are kept fixed.

After the first 200 iterations, both our action-dependent rules remain above the SAM~2 and SAM~3 rules, showing that the object pointer selection rule affects tracking performance.
Between the two, the recently evicted rule scores higher in most evaluations, but the longest-kept rule performs comparably and their seed bands overlap.
Designing better rules could improve performance further.
Moreover, since learned memory control outperforms hand-crafted rules for memory bank selection, extending the policy to also select object pointers is a natural direction for future work.

\subsection{Reward Normalization}
\label{abl-reward}

We train SAM3RL against the normalized reward $r_t = q_t - b_t$, which is the difference between the tracking quality $q_t$ of SAM~3 under our learned memory control and the tracking quality $b_t$ of SAM~3 under its native memory control.
Since much of the variance of $q_t$ comes from the difficulty of the video rather than from the actions taken, subtracting the policy-independent $b_t$ reduces this variance while leaving the optimal policy unchanged.
In \Cref{fig:ablation_reward}, we ablate this normalization by training on the raw quality $r_t = q_t$ with everything else unchanged, and evaluate both arms every 25 training iterations on the LaSOT validation split.
The raw-reward arm falls below SAM~3 itself on several evaluations in the first 250 iterations and is less stable than the difference-based arm, which stays above SAM~3 throughout.

\clearpage

\section{Extended analysis of learned memory control}
\label{suppl:analysis}

\begin{figure*}[htb]
\begin{minipage}[t]{0.49\textwidth}
\centering
\includegraphics[scale=0.98]{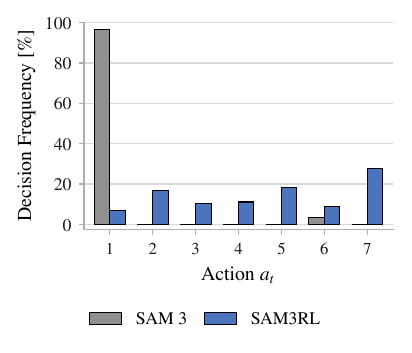}
\caption{\textbf{Action distributions.}
Frequency of each action $a_t$ for SAM~3 and SAM3RL on our LaSOT validation split.
Action $a_t \in \{1, \dots, 6\}$ evicts the stored memory $m_{\tau_{a_t}}$, where $a_t = 7$ rejects the incoming frame.
The SAM~3 rule uses only two actions.
It evicts the most recent stored frame $m_{\tau_6}$ ($a_t = 6$) if its effective IoU score falls below the memory threshold, and the oldest frame $m_{\tau_1}$ ($a_t = 1$) otherwise.}
\label{fig:actions}
\end{minipage}%
\hfill
\begin{minipage}[t]{0.49\textwidth}
\centering
\includegraphics[scale=0.98]{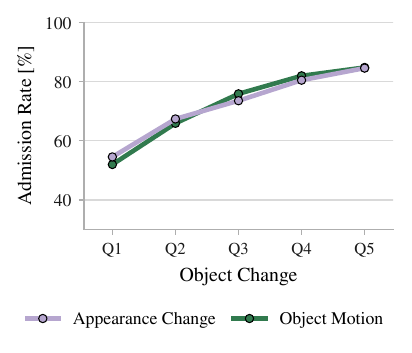}
\caption{\textbf{Admission follows object change.}
Share of incoming frames SAM3RL admits per quintile of object motion and appearance change on our LaSOT validation split.
\emph{Object motion} is the shift of the ground-truth box center between consecutive decisions, normalized by the square root of the box area.
\emph{Appearance change} is the DINOv3~\citep{simeoni2025dinov3} cosine distance between consecutive ground-truth crops.
}
\label{fig:cadence}
\end{minipage}
\end{figure*}

\begin{figure}[htb]
  \centering
  \includegraphics[scale=1]{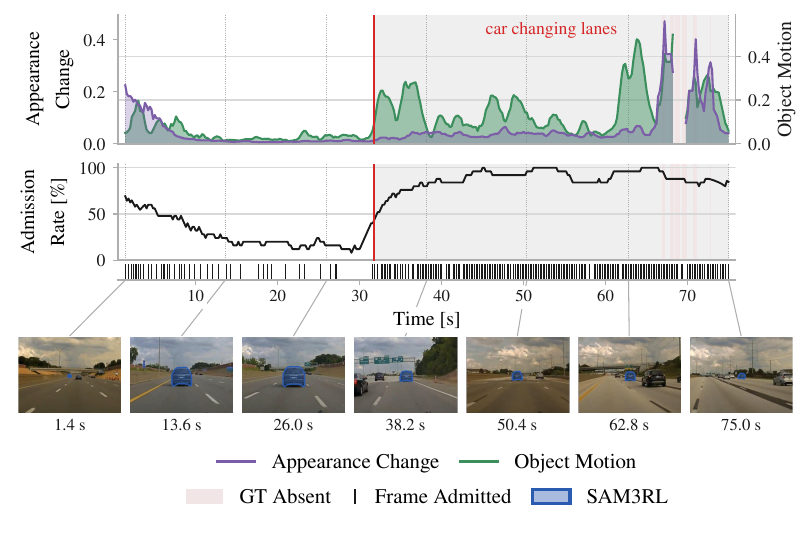}  
  \caption{\textbf{Admission rate of incoming frames by SAM3RL follows object change II.} A 75\,s DiDi video sequence in which a car cruises steadily in one lane and then starts moving quickly, changing lanes until the end. \emph{Object motion} is the shift of the ground-truth box center between consecutive decisions, normalized by the square root of the box area. \emph{Appearance change} is the DINOv3~\citep{simeoni2025dinov3} cosine distance between consecutive ground-truth crops. The middle panel is the admission rate in a 5\,s window, one tick per admitted decision. Admissions are sparse during the cruise and dense in the lane-changing interval (gray) when object motion rises.}
  \label{fig:car}
\end{figure}

We present the action distributions of SAM~3 and SAM3RL in \Cref{fig:actions}. SAM~3 evicts the oldest frame on almost every decision, while SAM3RL spreads its decisions over all seven actions, with no single action dominating. The admission rate of SAM3RL as a function of object motion and appearance change is shown in \Cref{fig:cadence}, where the admission rate rises both with object motion and object appearance change. This finding is consistent with the sequences shown in \Cref{fig:car,fig:rabbit}.

\begin{figure}[htb]
\centering
\includegraphics[scale=0.795]{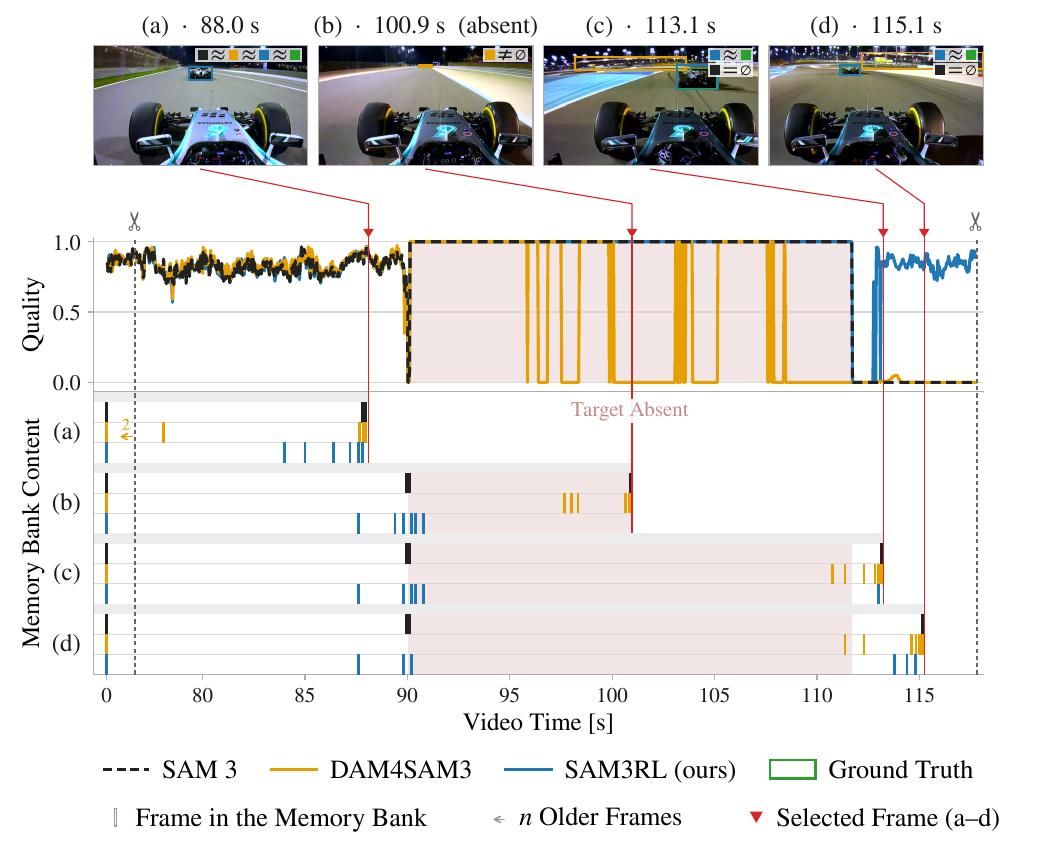}
\caption{\textbf{SAM3RL holds diverse pre-absence frames through a long absence.}
A VOT-LT$_{2021}$ sequence captured from the onboard camera of a formula car,
in which the car ahead leaves the field of view for about 22\,s. The top row
shows four selected frames (a--d) with the ground truth and the prediction of
each method. The middle panel shows the per-frame tracking quality of each
method in the selected time window, with the absence shaded red. The bottom
panel shows the memory bank content of each method at the four selected frames,
one row per selected frame and method and one tick per stored frame.
The prompt frame at 0\,s is on the left, and an arrow gives the number of stored
frames older than the selected window.
SAM~3 correctly predicts no target throughout the absence, with its bank fixed
on consecutive frames just before it, in which the car is mostly out of view.
It does not reacquire the car after its return.
DAM4SAM3 wrongly predicts the target at several points during the absence, and
by (b) its bank holds only frames from the absence. After the return, its
predictions miss the car.
SAM3RL also predicts no target during the absence. When the car leaves, its bank
holds frames spread over the last few seconds before the absence and a few from
the start of the absence, and at (b) and (c) it still holds nearly all of them.
SAM3RL reacquires the car about 1\,s after its return, and by (d) half of its
bank holds frames from after the return, while the pre-absence frames remain.}
\label{fig:case_f1}
\end{figure}

\begin{figure}[htb]
  \centering
  \includegraphics[scale=0.795]{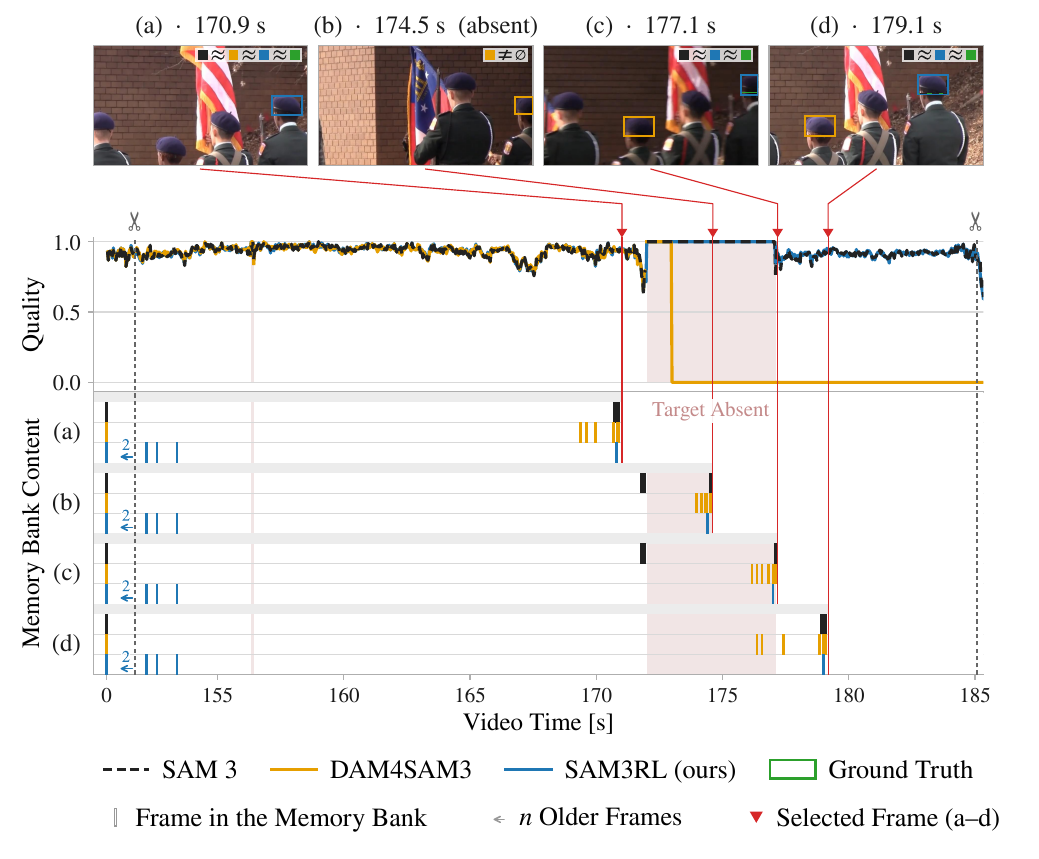}
  \caption{\textbf{SAM3RL holds onto very old frames and successfully reacquires.}
A LaSOT test sequence of a military ceremony, in which the tracked soldier's hat
leaves the field of view for about 5\,s among other soldiers wearing near-identical
hats. The top row shows four selected frames (a--d) with the ground truth and the
prediction of each method. The middle panel shows the per-frame tracking quality of
each method in the selected time window, with the absence shaded red. The bottom panel
shows the memory bank content of each method at the four selected frames, one row per
selected frame and method and one tick per stored frame. The prompt frame at 0\,s is
on the left, and an arrow gives the number of stored frames older than the selected
window. SAM~3 correctly predicts no target throughout the absence, with its bank holding
the last five frames before it, and reacquires the hat on the first frame after its
return. DAM4SAM3 also predicts no target at the start of the absence, but shortly after
it locks onto another soldier's hat.
By (b) it has evicted all pre-absence frames,
and its bank holds only frames in which it tracks the wrong hat. It keeps following that
hat for the rest of the sequence, also after the target returns. 
SAM3RL correctly predicts no target throughout the absence, but its bank barely changes. When
the hat leaves the view, five of its six slots hold frames that are 20--30\,s old. These
stay in the bank through more than 30\,s of tracking, from before (a) until after (d),
and only the newest slot turns over. SAM3RL thus keeps the target's
appearance in memory throughout the gap and reacquires the hat on the first frame after
its return.
}

  \label{fig:case_hat}
\end{figure}

\begin{figure}[htb]
  \centering
  \includegraphics[scale=0.795]{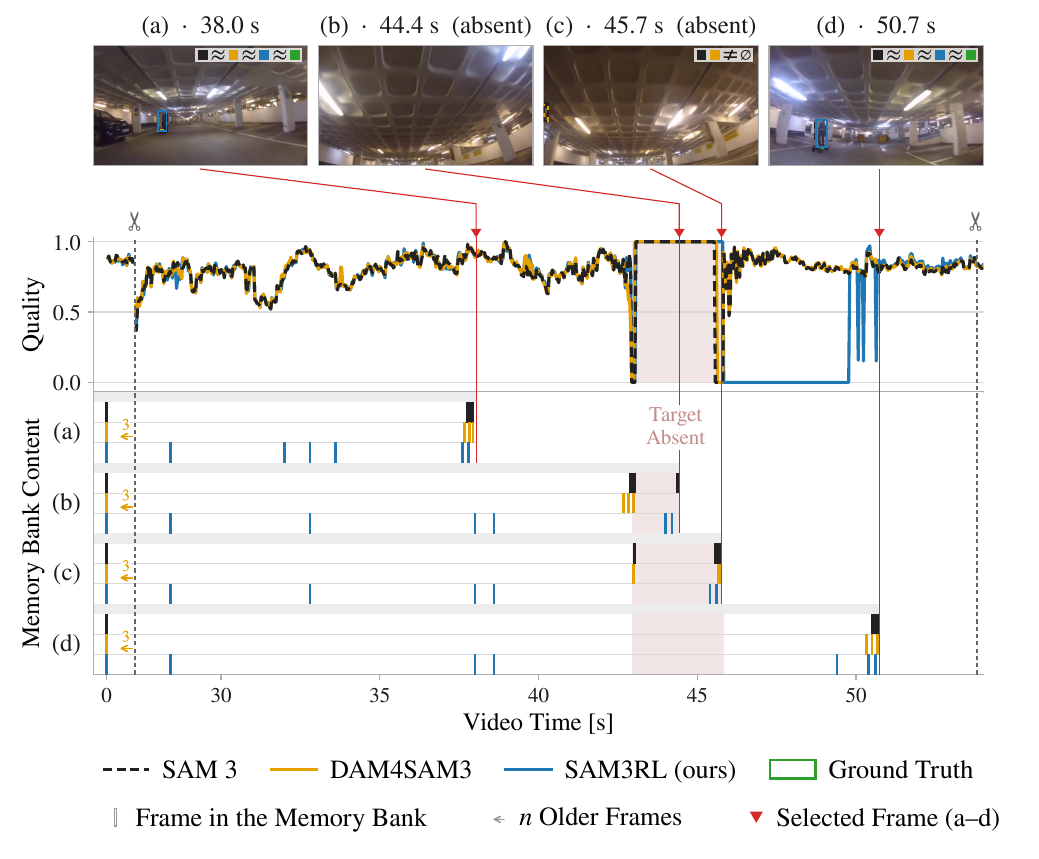}
  \caption{\textbf{SAM3RL holds sparse old frames and reacquires late.}
  A VOT-LT$_{2021}$ sequence capturing a longboarder from behind, who goes out of view on the left for about 3\,s. The top row
shows four selected frames (a--d) with the ground truth and the prediction of
each method. The middle panel shows the per-frame tracking quality of each
method in the selected time window, with the absence shaded red. The bottom
panel shows the memory bank content of each method at the four selected frames,
one row per selected frame and method and one tick per stored frame.
The prompt frame at 0\,s is on the left, and an arrow gives the number of stored
frames older than the selected window. SAM~3 correctly reports the target absent on most
frames of the gap and reacquires successfully. DAM4SAM3 keeps 3 old frames and 3 recent ones in all selected frames, although during the absence the 3 recent ones are the pre-absence ones. SAM3RL carries four frames from before the absence through the gap, which are 4--15\,s old at the start of the absence and all show the rider small and near the center of the frame, while the rider comes back large and on the left side. The memory bank frames that would match that view all passed through the two newest slots and are not present in the bank of SAM3RL at the end of the absence, but are present in the banks of SAM~3 and DAM4SAM3. SAM3RL fails to reacquire for 4\,s after the return, while SAM~3 and DAM4SAM3 reacquire immediately. Once the rider moves back towards the center, SAM3RL reacquires and updates its memory bank.}
  \label{fig:case_longboard}
\end{figure}

\Cref{fig:case_f1,fig:case_hat,fig:case_longboard} compare the memory banks of SAM~3, DAM4SAM3, and SAM3RL around three absence gaps.
In all three, SAM3RL keeps old, temporally sparse frames and admits few new ones while the target is absent, so the pre-absence appearance is still in its bank when the target returns.
In \Cref{fig:case_f1}, this lets SAM3RL reacquire the formula car that both SAM~3 and DAM4SAM3 lose.
In \Cref{fig:case_hat}, SAM3RL reacquires the soldier's hat, which DAM4SAM3 swaps for a distractor.
On the other hand, in \Cref{fig:case_longboard}, its bank retains frames showing the longboarder small and centered, but the rider returns large and on the left, so SAM3RL reacquires later than SAM~3 and DAM4SAM3.

\end{document}